\documentclass[10pt,twocolumn,letterpaper]{article}

\usepackage[pagenumbers]{cvpr} %

\usepackage{wrapfig}
\usepackage{bbm}
\usepackage{amssymb}
\usepackage{pifont}
\newcommand{\cmark}{\ding{51}} %
\newcommand{\xmark}{\ding{55}} %

\graphicspath{{./}}

\usepackage{amsmath}
\providecommand{\augset}{\mathcal{A}}
\usepackage{multirow}

\PassOptionsToPackage{table}{xcolor}
\usepackage{acro}
\acsetup{make-links=false}

\usepackage{amsmath,amsfonts,bm}

\def\eqref#1{equation~\ref{#1}}

\def\1{\bm{1}}

\newcommand{\dset}{\mathcal{X}}               %
\newcommand{\simg}{\hat{x}}                   %

\newcommand{\enc}{E}                          %

\newcommand{\idl}{\mathrm{id}}                %
\newcommand{\reid}{R}                         %

\DeclareRobustCommand{\AUCROC}{\ensuremath{\text{AUC-ROC}}\xspace}

\input{acronyms/acronym-symbol.ac}
\input{acronyms/acronym-abbreviation.ac}
\usepackage{colortbl}
\definecolor{fau}{HTML}{04316A}
\definecolor{fau_tech}{HTML}{8C9FB1}
\definecolor{fau_phil}{HTML}{FDB735}
\definecolor{fau_rw}{HTML}{C50F3C}
\definecolor{fau_med}{HTML}{18B4F1}
\definecolor{fau_nat}{HTML}{7BB725}

\definecolor{fau_alpha1}{HTML}{617DA1}
\definecolor{fau_alpha2}{HTML}{A0B1C6}
\definecolor{fau_alpha3}{HTML}{C0CBDA}

\definecolor{fau_tech_alpha1}{HTML}{B6C2CE}
\definecolor{fau_tech_alpha2}{HTML}{D3DAE1}
\definecolor{fau_tech_alpha3}{HTML}{E2E7EB}

\definecolor{fau_nat_alpha1}{HTML}{ACD275}
\definecolor{fau_nat_alpha2}{HTML}{CDE4AC}
\definecolor{fau_nat_alpha3}{HTML}{DEEDC8}

\definecolor{fau_med_alpha1}{HTML}{6DD0F6}
\definecolor{fau_med_alpha2}{HTML}{A7E2FA}
\definecolor{fau_med_alpha3}{HTML}{C5ECFB}

\definecolor{fau_rw_alpha1}{HTML}{DA6784}
\definecolor{fau_rw_alpha2}{HTML}{E9A4B5}
\definecolor{fau_rw_alpha3}{HTML}{F0C3CE}

\definecolor{fau_phil_alpha1}{HTML}{FECE76}
\definecolor{fau_phil_alpha2}{HTML}{FEE4B2}
\definecolor{fau_phil_alpha3}{HTML}{FEEDCC}

\definecolor{fau_light}{HTML}{CED9E7}
\definecolor{fau_light_tech}{HTML}{D3DDE6}
\definecolor{fau_light_phil}{HTML}{F4E8CB}
\definecolor{fau_light_rw}{HTML}{F2DED1}
\definecolor{fau_light_med}{HTML}{B9D9EC}
\definecolor{fau_light_nat}{HTML}{CFE0D8}

\definecolor{fau_dark}{HTML}{041E42}
\definecolor{fau_dark_tech}{HTML}{2F586E}
\definecolor{fau_dark_phil}{HTML}{FDB735}
\definecolor{fau_dark_rw}{HTML}{971B2F}
\definecolor{fau_dark_med}{HTML}{005287}
\definecolor{fau_dark_nat}{HTML}{266141}

\definecolor{ecblue}{HTML}{4C72B0}
\definecolor{emorange}{HTML}{DD8452}

\iftoggle{cvprfinal}{}{%
  \definecolor{fau}{gray}{0.25}%
  \definecolor{fau_tech_alpha3}{gray}{0.90}%
}

\definecolor{cvprblue}{rgb}{0.21,0.49,0.74}
\usepackage[pagebackref,breaklinks,colorlinks,allcolors=cvprblue]{hyperref}

\title{LCMem: Latent Consistent Image Memorization Detection}

\author{Mischa Dombrowski$^{1}$\quad
Felix N\"utzel$^{1}$\quad
Bernhard Kainz$^{1,2}$\\
$^{1}$Friedrich-Alexander-Universit\"at Erlangen-N\"urnberg\quad
$^{2}$Imperial College London
}

\begin{document}
\def\lcmemteaserontitlepage{}
\twocolumn[{%
    \renewcommand\twocolumn[1][]{#1}%
    \maketitle
    \begin{center}
        \newcommand{\lcmemteasercaption}[2][]{\captionof{figure}[#1]{#2}}
        \centering
\includegraphics[width=\linewidth]{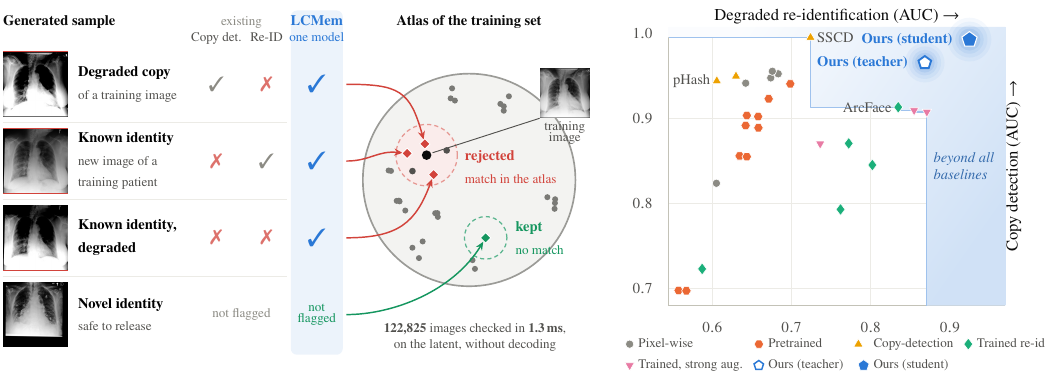}
\lcmemteasercaption[LCMem: one check against the training-set atlas]{\textbf{Left:} A generated sample can leak training data as a degraded copy of a training image or as a new, possibly degraded image of a training identity. Copy detectors and re-identification models each catch only one of these cases. LCMem is a single embedding model that scores the latent of a sample against an atlas of the entire training set and rejects all three, while novel identities are kept (schematic). \textbf{Right:} Each point is one method (AUC at corruption strength 4, mean over five datasets and six corruptions); our student is the only method that is strong on both axes. The line is the Pareto front of the baselines; no baseline reaches the shaded area beyond it.}
\label{lcmem:fig:abstract}

    \end{center}
    \vspace{0.5em}
}]
\begin{abstract}
Generative image models may reproduce training data, so they cannot be used for privacy-protecting data sharing without infeasible human checks.
Unfortunately, current memorization detectors are incomplete. Copy detection only recognizes degraded duplicates of a training image, but misses different images of the same person, while 
re-identification recognizes the person, but fails once the image is degraded.
We introduce the Latent Consistent Memorization Network (LCMem), a single embedding model, unifying copy detection and re-identification directly in the latent spaces of generative models.
A teacher acquires identities jointly on the latents of three autoencoders. A noise-conditioned student is distilled from it under image augmentation and latent noise. This enables one model to score both the clean images and noisy intermediate states of a diffusion sampler.
LCMem detects $69.6\,\%$ of the degraded copies and $20.6\,\%$ of the degraded images of a known identity across five datasets, against $56.9\,\%$ and $6.3\,\%$ for the best baseline.
In deployment, LCMem compares a sample with the entire training set of $122{,}825$ images in $1.3$\,ms without decoding. It separates training identities from novel ones with an \AUCROC{} of $0.996$ (teacher) and $0.944$ (student).
Thus, a privacy check against the full training corpus  becomes cheap enough to run at every sampling step and  as part of a training loss.
Code will be made available upon publication.
\end{abstract}

\section{Introduction}
\ifdefined\lcmemteaserontitlepage\else
\begin{figure*}[t]
    \let\lcmemteasercaption\caption
    \IfFileExists{sec/fig_teaser.tex}{\input{sec/fig_teaser}}{\input{sec/fig_teaser}}
\end{figure*}
\fi

Synthetic data sharing has emerged as a promising avenue to overcome long-standing barriers to data access and collaboration.
In healthcare, the majority of medical images are confined to isolated databases and electronic health record systems, where privacy concerns and regulatory constraints prevent data exchange across institutions~\cite{rieke_future_2020,ng_federated_2021,white_data_2020,yahiaoui_federated_2024,asiimwe_biobank_2021,saberi_data_2025,nissim_privacy_2021}.
Legal frameworks such as the EU General Data Protection Regulation (GDPR) further restrict the processing of personal data, requiring that identifiability be excluded by ``all means reasonably likely to be used''~\cite{vollmer_recital_2023,dombrowski_bridge}.
In the natural image domain, copyright and data-ownership laws confine high-value datasets 
within proprietary repositories, often prohibiting redistribution or reuse for machine-learning purposes~\cite{kretschmer_copyright_2024,buick_copyright_2025,margoni2022deeper}.

Compared with federated learning~\cite{mcmahan_communication-efficient_2017,kairouz_advances_2021,zhu_deep_2019} and dataset distillation~\cite{welling_herding_2009,yu_dataset_2023}, synthetic data generation is a direct, infrastructure-agnostic route to shareable datasets that preserve the statistics of the original data~\cite{reynaud2024echonet,dombrowski_cvpr,dombrowski_bridge}.

A central limitation of synthetic data sharing is the need for reliable mechanisms that guarantee the generated samples are truly synthetic and not memorized replicas of real images.
The prevailing strategy is to train a model that predicts whether two images correspond to the same underlying instance, either through re-identification~\cite{packhäuser2022deeplearning} or self-supervised similarity learning~\cite{reynaud2024echonet,molina_memorization_2024,dar2024unconditional}, yet these approaches rely on global dataset-level statistics, which leads to inconsistent behavior and overconfident decisions~\cite{koeken2025sensitivityspecificity}.
They are typically evaluated on a single dataset, and they treat re-identification, whether two images depict the same identity, and copy detection, whether an augmented image still originates from the same source, as separate problems.
For reliable privacy auditing, these tasks must be addressed jointly and evaluated across datasets and modalities, as their performance diverges markedly (see \cref{lcmem:fig:abstract}).

Our contributions are fourfold.
\begin{itemize}
    \item[(i)] We propose memorization detection as a unified problem that spans copy detection and re-identification: a memorized sample can be a degraded copy of a training image or a different image of the same identity, and a detector has to recognize both (\cref{lcmem:sec:problem}).
    \item[(ii)] We introduce LCMem, a memorization detector that operates in generative latent spaces and is trained in two stages, a teacher and a noise-conditioned student. At a threshold that rejects $5\,\%$ of novel images, the student has the highest recall of all methods on copy detection and degraded re-identification, and the teacher on re-identification (\cref{lcmem:tab:main}).
    \item[(iii)] We show that LCMem is usable as a one-vs-all identity check against the full training set. On the training split, which is the set a privacy filter has to protect,
 LCMem separates known from novel identities with an \AUCROC{} of up to $0.996$ across five datasets, at $1.3$\,ms per sample against up to $122{,}825$ images 
    (\cref{lcmem:sec:ova_deployment}).
    \item[(iv)] We show that LCMem can be applied to the noisy intermediate latents of a diffusion sampler: the identity of a sample is fixed and readable from step 20 of 32 on (\cref{lcmem:sec:applied}).  This moves the privacy check from the finished image into the sampling loop. %
\end{itemize}

\section{Related Work}
\noindent\textbf{Image Generation:}
Diffusion models~\cite{song2020denoising,ho2020denoising} are the dominant paradigm for high-fidelity image synthesis, and latent diffusion models~\cite{rombach2022high} scale them by operating in the compressed latent space of an autoencoder, typically $64 \times 64 \times 4$ for a $512 \times 512 \times 3$ image.
Diversity-aware diffusion models~\cite{dombrowski_cvpr} and PSO-secure generative models~\cite{dombrowski_bridge} extend them toward privacy-preserving synthesis, in which no generated sample can be linked to a specific individual or record.

\noindent\textbf{Memorization:}
Memorization has emerged as a critical vulnerability in generative modeling, as models can unintentionally reproduce training data with near-identical fidelity, revealing private or copyrighted content \cite{dar2024unconditional,carlini2023extracting}.
Approaches to address this issue broadly divide into those that intervene during learning or sampling, and post-hoc filtering techniques applied after generation.
The first group modifies the model's optimization or sampling dynamics using mechanisms, like cross-attention regularization, data augmentation, or privacy-preserving objectives, to constrain overfitting~\cite{ren2024unveiling,dar2024unconditional}.
Differential privacy represents the strongest formal guarantee~\cite{ziller_reconciling_2024,zhao2025does},
but is impractical for high-dimensional or high-resolution image synthesis, where utility degradation is severe and scalability limited.
Post-hoc filtering, in contrast, does not alter model behavior but instead identifies and removes memorized outputs after sampling using re-identification, similarity search, or membership-inference based attacks~\cite{packhäuser2022deeplearning,dar_investigating_2023,reynaud2024echonet,carlini_membership_2022,carlini2023extracting,dombrowski_bridge,kriplani2025solidmarkevaluating}.
However, as shown by recent large-scale extraction studies \cite{koeken2025sensitivityspecificity}, existing filters are still limited in specificity and consistency, often failing to detect reconstructive or partial memorization. 
Existing filters use either a copy detector or a re-identification model, and neither family covers the failure cases of the other. 

\noindent\textbf{Copy Detection and Re-Identification:}
Two lines of work provide the detectors that post-hoc filtering relies on.
\emph{Copy detection} asks whether an image is an edited or degraded version of a reference image.
Classical methods compare perceptual hashes such as pHash~\cite{zaunerimplementationbenchmarking}; learned descriptors such as SSCD~\cite{pizzi2022selfsuperviseddescriptor} are trained with self-supervised contrastive objectives and strong augmentation on DISC21~\cite{douze20222021image} and have become the common tool to measure replication in diffusion models~\cite{somepalli2023understandingmitigating}.
Shi et al.~\cite{shi2026privacypreservinglatent} compare the features of a pretrained classification network.
These methods are robust to corruptions by construction but have no notion of identity: a different image of the same patient is not a copy.
\emph{Re-identification} asks whether two different images show the same identity.
Siamese networks trained on patient labels re-identify patients across examinations~\cite{packhäuser2022deeplearning}, contrastive variants are used as privacy filters for synthetic medical images~\cite{reynaud2024echonet,dar2024unconditional}, and margin-based identity classification such as ArcFace~\cite{deng2019arcface} is the standard in face recognition.
These models are trained on clean images and lose their accuracy once the query is degraded.
Both lines operate on finished images in pixel space.
Classifiers conditioned on the noise level are known from classifier guidance~\cite{dhariwal2021diffusion}, but we are not aware of a memorization detector that works on the noisy latents of a sampler.
LCMem joins the two lines in one embedding and moves it into the latent space of the generator.

\section{Method}
\begin{figure*}[t]
    \centering
    \includegraphics[width=\linewidth]{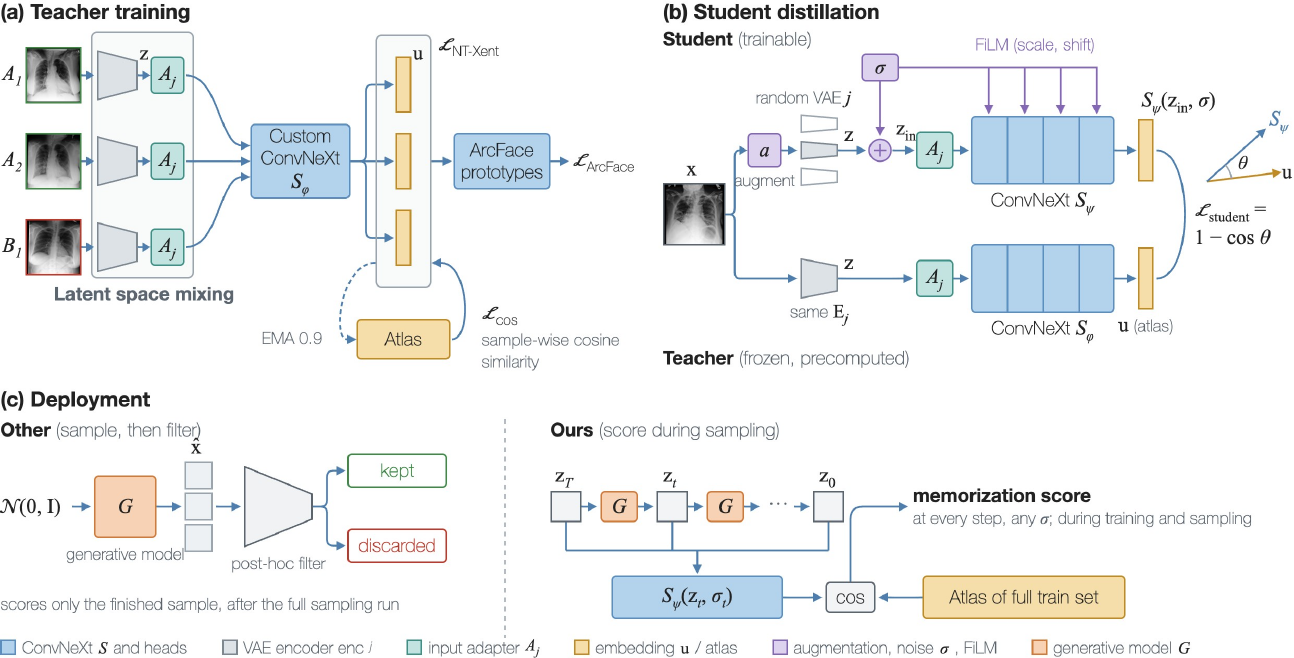}
    \caption[LCMem overview: teacher, student and deployment]{LCMem overview.
    \textbf{(a) Teacher.} Clean latents of image pairs, each from an independently drawn autoencoder $\enc_j$ with its adapter $A_j$, are embedded by one ConvNeXt $S_\phi$ and trained with a contrastive, an atlas and an identity loss (\cref{lcmem:sec:teacher}).
    \textbf{(b) Student.} A copy of the teacher with FiLM conditioning on the noise level $\sigma$ maps an augmented and noised latent to the frozen teacher's embedding of the clean image (\cref{lcmem:sec:student}).
    \textbf{(c) Deployment.} Existing filters score the finished sample $\simg$; the student scores any intermediate state $z_t$ at its noise level $\sigma_t$ against the atlas of the training set.}
    \label{lcmem:fig:overview}
\end{figure*}
\subsection{Problem Formulation}
\label{lcmem:sec:problem}
Let $\dset$ be a dataset of images and $\idl(x)$ the identity label of $x \in \dset$: the patient in medical datasets, or a person, creator or stylistic identifier in natural image datasets.
A memorization detector is a model $\reid(x, x')$ that scores whether two images share an identity and flags the pair as memorized if $\reid(x, x') > \delta$ for a threshold $\delta$.

Let $\augset$ be a set of image corruptions $a$ (blur, noise, rotation, brightness, contrast, JPEG compression, each at several strengths).
Depending on whether the second image is the same image and whether it is corrupted, there are three settings with the positive pairs
\begin{align}
\mathcal P^{+}_{\mathrm{copy}} &= \{(x, a(x)) : a \in \augset\}, \\
\mathcal P^{+}_{\mathrm{reid}} &= \{(x, x') : \idl(x) = \idl(x'),\, x \neq x'\}, \\
\mathcal P^{+}_{\mathrm{deg}} &= \{(x, a(x')) : \idl(x) = \idl(x'),\, x \neq x',\, a \in \augset\}.
\end{align}
\emph{Copy detection} asks whether a corrupted image still originates from the same source, \emph{re-identification} whether a different image shows the same identity, and \emph{degraded re-identification} combines both.
The negative pairs $\mathcal P^{-}$ of each setting consist of images with different identities, $\idl(x) \neq \idl(x')$, corrupted in the same way as the positives.
Copy detection and re-identification are the special cases $x' = x$ and $a = \mathrm{id}$ of degraded re-identification, so a single model $\reid$ has to solve all three.
At deployment time, a generated image $\simg$ is memorized if its best match in the training set exceeds the threshold,
\begin{equation}
\bar{\reid}(\simg) = \max_{x \in \dset} \reid(x, \simg) > \delta .
\end{equation}

\subsection{Latent Embedding Network}
\label{lcmem:sec:latent}
\label{lcmem:sec:network}
Instead of operating in pixel space, LCMem encodes each image with the autoencoder of a generative model.
The encoder discards imperceptible high-frequency detail that carries little identity information, and at sampling time the generator synthesizes the latent, so no decoding is needed.
For an image $x$ and autoencoder $j$ with encoder $\enc_j$, we use the posterior mean $z^{(j)} = \enc_j(x)$.
We train with the SD2 autoencoder \cite{rombach2022high}, the SDXL autoencoder~\cite{podell2024sdxlimproving}, and the FLUX.1 autoencoder~\cite{blackforestlabs_flux1}.
To test generalization, we add the unseen FLUX.2 autoencoder~\cite{blackforestlabs_flux2} after training by training a new adapter only, with all other weights frozen (\cref{lcmem:sec:new_autoencoders}).

LCMem is a single embedding network $S_\phi$ that maps a latent to a vector $u = S_\phi(z) \in \mathbb{R}^{512}$ (\cref{lcmem:fig:overview}).
Both images of a pair are embedded independently with the same weights, and the pair score is the cosine similarity
\begin{equation}
\reid(x, x') = \mathrm{sim}(u, u') = \frac{u^\top u'}{\lVert u \rVert \, \lVert u' \rVert} .
\end{equation}
Latent shapes differ across autoencoders (4 to 32 channels), so each latent is scaled by a constant $\kappa_j$ per autoencoder and mapped by a small convolutional adapter $A_j$ to a common width of $32$ channels; everything after the adapter is shared.
The backbone is a ConvNeXt-Tiny~\cite{liu2022convnext} whose first stride-4 stem is replaced by a $1{\times}1$ convolution with stride 1, so that the first stage runs on the full $64{\times}64$ latent grid.
We show that a stem with stride 4 or stride 2 lowers the re-identification \AUCROC{} in \cref{lcmem:tab:ablation}.
The backbone and the adapters are initialised by feature distillation from a ConvNeXt-Large on ImageNet.
Layer-level details of the network and of both training stages are in \cref{lcmem:sec:impl_details}.

\subsection{Teacher}
\label{lcmem:sec:teacher}
The teacher (\cref{lcmem:fig:overview}a) is trained on clean, precomputed latents only, without any augmentation or noise.
Its purpose is identity discrimination and the alignment of the three latent spaces to one embedding space.

\noindent\textbf{Training data.}
We train on six datasets $\mathcal{D} = \{$CelebA, CT-RATE, NIH-CXR-LT, ImageNet, ISIC, MIMIC-CXR$\}$ jointly (\cref{lcmem:tab:dataset_overview}).
Each training sample is a pair $(x, x')$ from one dataset, with the dataset drawn uniformly from $\mathcal{D}$.
With probability $0.5$ the pair is positive and $x'$ is drawn from the same identity; otherwise $x'$ is drawn from another identity of the same dataset.
The autoencoder is drawn independently and uniformly for $x$ and for $x'$, so two thirds of the pairs mix latent spaces.

\noindent\textbf{Loss.}
Three terms act on the same embedding $u$ at three granularities.
\begin{equation}
\mathcal{L}_{\mathrm{teacher}}
= \lambda_{\mathrm{nce}} \mathcal{L}_{\mathrm{nce}}
+ \lambda_{\mathrm{atlas}} \mathcal{L}_{\mathrm{atlas}}
+ \lambda_{\mathrm{arc}} \mathcal{L}_{\mathrm{arc}} ,
\end{equation}
with $\lambda_{\mathrm{nce}} = 0.5$, $\lambda_{\mathrm{atlas}} = 1$ and $\lambda_{\mathrm{arc}} = 1$.

\emph{Contrastive term (batch).}
$\mathcal{L}_{\mathrm{nce}}$ is the NT-Xent loss~\cite{chen2020simpleframework} with temperature $\tau = 0.07$ over the positive pairs of the batch, where the embeddings of the other positive pairs are the negatives.
This is the only term that compares two embeddings directly.

\emph{Atlas term (image).}
Every training image $x_i$ owns one entry $\bar{u}_i \in \mathbb{R}^{512}$ in a memory, the training-time atlas, which is the exponential moving average of its unit-length embedding over all its previous visits, in whichever latent space the visit used.
Such a per-instance memory with a momentum update follows instance discrimination~\cite{wu2018unsupervisedfeature} and the momentum principle of MoCo~\cite{he2020momentumcontrast}.
For every image of the batch that already has an entry, the loss is the cosine distance to that entry,
\begin{equation}
\mathcal{L}_{\mathrm{atlas}} = \frac{1}{|\mathcal{B}|} \sum_{i \in \mathcal{B}} \big( 1 - \mathrm{sim}(u_i, \bar{u}_i) \big),
\end{equation}
where $\mathcal{B}$ is the set of batch images with an entry and $\bar{u}_i$ is treated as a constant (no gradient flows into the atlas).
Because the index is the image and not the (image, autoencoder) pair, the SD2, SDXL and FLUX.1 encodings of one image are all pulled toward the same point. This aligns the three latent spaces.

\emph{Identity term (identity).}
For every dataset with identity labels (all but ImageNet) we keep a separate ArcFace head~\cite{deng2019arcface} with one learned prototype per training identity.
$\mathcal{L}_{\mathrm{arc}}$ is the standard ArcFace loss with the identity $\idl(x)$ of every batch image as label; the softmax runs over the identities of one dataset only, so identities of different datasets never compete.
The prototypes are used for training only and discarded afterwards.

\subsection{Student}
\label{lcmem:sec:student}
The student $S_\psi$ (\cref{lcmem:fig:overview}b) is the deployed model.
The teacher only sees clean latents, but a latent diffusion model produces an image by denoising: its intermediate states $z_t$ are noisy latents, and the detector should be able to score them before the image is finished.
The student is therefore the teacher with an additional noise-level conditioning, fine-tuned so that an augmented and noised view of an image is embedded where the teacher embeds the clean image.
It learns no identities and sees no negatives; all discrimination is inherited from the teacher.

\noindent\textbf{Noise-level conditioning.}
Like a denoiser, the detector faces inputs of all noise levels, and the sampler knows the level $\sigma_t$ of its state.
We therefore give $\sigma$ to the network as a second input and let it adapt its features with FiLM~\cite{perez2017filmvisualreasoninggeneral}: a per-channel scale and shift, computed from an embedding of $\log\sigma$ and applied after each of the four ConvNeXt stages.

\noindent\textbf{Training.}
Each training image is augmented in pixel space (rotation, flip, blur, noise and JPEG compression; \cref{lcmem:sec:impl_details}) and encoded by the frozen encoder $\enc_j$ of a uniformly drawn autoencoder.
The scaled latent is then noised with
\begin{equation}
z_{\mathrm{in}} = \frac{\kappa_j z^{(j)} + \sigma \epsilon}{\sqrt{1 + \sigma^2}}, \qquad \epsilon \sim \mathcal{N}(0, \mathbf{I}),
\end{equation}
which keeps the input variance independent of $\sigma$, with $\log\sigma \sim p_\sigma = U[-4, 3]$, except that with probability $0.2$ the latent is left clean ($\sigma = 0$).
The target is the frozen teacher's embedding $u^{(j)}(x)$ of the clean, un-augmented image in the same latent space, and the loss is the cosine distance
\begin{equation}
\mathcal{L}_{\mathrm{student}} = 1 - \mathrm{sim}\big( S_\psi(z_{\mathrm{in}}, \sigma),\; u^{(j)}(x) \big) .
\end{equation}
No identity label, no second image and no negative is involved: the student only learns invariance to the pixel augmentation and to the latent noise, and the dependence on $\sigma$.
At test time the student is used with $\sigma = 0$ for images (\cref{lcmem:fig:overview}c) and with the noise level of the sampler state for noisy latents (\cref{lcmem:sec:applied}).

\section{Experiments}

\noindent\textbf{Datasets:} We evaluate on CelebA, MIMIC-CXR, NIH-CXR-LT, ISIC-2020 and CT-RATE, all with identity labels, and use ImageNet as additional training data only (\cref{lcmem:tab:dataset_overview,lcmem:sec:datasets}).

\noindent\textbf{Evaluation protocol:}
\label{lcmem:sec:protocol}
We evaluate the three settings of \cref{lcmem:sec:problem} on the test split of each dataset, using the same pairs, corruptions and threshold rule for every method.
All images are resized to $512 \times 512$ RGB.
Per dataset we draw 1000 positive pairs $(x, x')$ of different images with the same identity, such that every possible pair is equally likely.
These pairs form the re-identification setting, the same pairs with a corrupted second image $a(x')$ form degraded re-identification, and copy detection uses $(x, a(x))$ for the anchors $x$ of these pairs.
The negatives are 1000 pairs of random test images with different identities, corrupted with the same $a$ as the positives.
$\augset$ consists of six transforms (Gaussian blur, Gaussian noise, rotation, brightness, contrast, JPEG compression) with four strengths each (\cref{lcmem:sec:protocol_details}), applied to the second image only and seeded, so all methods see identical corrupted images.

\noindent\textbf{Metrics:}
For every setting we report the threshold-free \AUCROC{} between positives and negatives and the recall of the positive pairs at a one-vs-all threshold $\delta_\rho$ with $\rho = 5\,\%$ (R@5), both averaged over datasets (and over corruptions for copy detection and degraded re-identification).
The threshold follows the deployment of \cref{lcmem:sec:problem}, where a filter rejects a generated image $\simg$ if $\bar{\reid}(\simg) > \delta$.
We fit one threshold per method and dataset on the validation split and use it unchanged in all three settings: $\delta_\rho$ is the $(1-\rho)$ quantile of $\bar{\reid}(\simg)$ over the queries, so the filter rejects a fraction $\rho$ of novel images (image-level false positive rate).
$\rho$ is the oversampling cost of the filter: a rejection-sampling loop has to draw $1/(1-\rho)$ samples per accepted novel image.
Since the threshold is one-vs-all, it depends on the size of the training set it was fitted against.

\noindent\textbf{Benchmark models:}
We compare LCMem with 26 baselines in four groups (details in \cref{lcmem:sec:baseline_details}). \Cref{lcmem:tab:main} shows a representative subset; the full table is \cref{lcmem:tab:main_full} in the appendix.
\emph{Pixel-wise metrics:} the mean squared error (MSE), the mean absolute error (L1), the normalized cross-correlation (NCC) and \ac{SSIM} \cite{wang_image_2004} at $512 \times 512$.
\emph{Pretrained encoders:} the cosine similarity of the features of BYOL \cite{grill_bootstrap_2020}, CLIP \cite{radford2021learning}, ConvNeXt \cite{liu2022convnext}, data2vec \cite{baevski_data2vec_2022}, DINOv2 \cite{oquab_dinov2_2024}, Inception \cite{szegedy_rethinking_2016}, MAE \cite{he_masked_2022} and SwAV \cite{caron_unsupervised_2020}, following \cite{stein2024exposing, dombrowski_cvpr}, extended with DINOv3 \cite{siméoni2025dinov3}, the VGG19 \cite{simonyan2014very} features of \cite{shi2026privacypreservinglatent} and an untrained Inception-v3 (``Random'') \cite{oareilly_pre-trained_2021}.
\emph{Copy detection:} SSCD~\cite{pizzi2022selfsuperviseddescriptor}, a common memorization detector for diffusion models~\cite{somepalli2023understandingmitigating}, and the perceptual hash pHash~\cite{zaunerimplementationbenchmarking}.
\emph{Trained on our data:} the Siamese re-identification model of \cite{packhäuser2022deeplearning}, the contrastive Siamese + NT-Xent of \cite{dar2024unconditional} with and without identity labels, and ArcFace~\cite{deng2019arcface} with a ResNet-101 and with a ConvNeXt-Tiny backbone, the image-space counterpart of the LCMem backbone.
They are trained on the training data of LCMem at $512 \times 512$; the rows marked \emph{strong aug.} replace the weak augmentation of the original methods by the pixel augmentation of the LCMem student and are the augmentation-matched control.
The threshold of every baseline is fitted with the same protocol as for LCMem.

\noindent\textbf{Training details:}
LCMem is trained as described in \cref{lcmem:sec:network,lcmem:sec:teacher,lcmem:sec:student}: the distillation of the initialisation takes $7.6$\,h on one A100, the teacher $7.7$\,h on one H100 and the student $12$\,h on one A100.
Ablations of the initialisation, the stem and the losses are in \cref{lcmem:sec:ablations}.

\begin{table*}[t]
    \centering\small
    \renewcommand{\arraystretch}{0.92}\setlength{\tabcolsep}{12pt}\footnotesize
    \resizebox{\textwidth}{!}{%
    \IfFileExists{figures/main_results_table_short.tex}{\begin{tabular}{llcccccc}
\toprule
\rowcolor{fau}
\color{white}\textbf{Type} & \color{white}\textbf{Method} & \multicolumn{2}{c}{\color{white}\textbf{Copy detection}} & \multicolumn{2}{c}{\color{white}\textbf{Re-identification}} & \multicolumn{2}{c}{\color{white}\textbf{Degraded re-id}} \\
\cmidrule(lr){3-4}\cmidrule(lr){5-6}\cmidrule(lr){7-8}
 &  & AUC & R@5 & AUC & R@5 & AUC & R@5 \\
\midrule
\multirow{3}{*}{Pixel-wise} & MSE & 0.948 & 33.5 & 0.703 & 2.9 & 0.674 & 1.0 \\
 & NCC & 0.956 & 56.2 & 0.698 & 2.5 & 0.676 & 1.6 \\
 & SSIM & 0.942 & 40.8 & 0.663 & 4.2 & 0.642 & 1.9 \\
\midrule
\multirow{5}{*}{Pretrained} & CLIP & 0.856 & 4.7 & 0.695 & 1.2 & 0.634 & 0.0 \\
 & SwAV & 0.941 & 15.5 & 0.763 & 3.3 & 0.698 & 0.1 \\
 & DINOv2 & 0.904 & 17.0 & 0.702 & 1.4 & 0.644 & 0.0 \\
 & DINOv3 & 0.924 & 19.4 & 0.741 & 1.5 & 0.671 & 0.1 \\
 & VGG19 & 0.892 & 3.3 & 0.710 & 1.3 & 0.642 & 0.0 \\
\midrule
\multirow{2}{*}{Copy det.} & pHash & 0.945 & 56.9 & 0.620 & 0.9 & 0.606 & 0.6 \\
 & SSCD & \textbf{0.996} & 48.7 & 0.749 & 2.8 & 0.724 & 0.4 \\
\midrule
\multirow{8}{*}{Trained} & Siamese & 0.846 & 0.4 & 0.948 & 7.0 & 0.802 & 0.2 \\
 & Siamese + NT-Xent & 0.793 & 0.6 & 0.969 & 5.2 & 0.762 & 0.0 \\
 & Siamese + NT-Xent (unsup.) & 0.723 & 1.1 & 0.708 & 0.5 & 0.587 & 0.0 \\
 & Siamese (strong aug.) & 0.910 & 0.5 & 0.945 & 2.3 & 0.855 & 0.1 \\
 & Siamese + NT-Xent (strong aug.) & 0.908 & 5.3 & \textbf{0.970} & 5.0 & 0.871 & 0.4 \\
 & ArcFace & 0.871 & 20.1 & 0.956 & 33.1 & 0.772 & 1.4 \\
 & ArcFace (ConvNeXt-T) & 0.914 & 43.0 & 0.963 & 44.4 & 0.835 & 6.3 \\
 & ArcFace (strong aug.) & 0.871 & 21.1 & 0.874 & 10.8 & 0.736 & 2.2 \\
\midrule[1.6pt]
\rowcolor{black!12}Ours & LCMem (teacher) & 0.966 & 48.1 & 0.947 & \textbf{49.6} & 0.868 & 13.4 \\
\rowcolor{black!12} & LCMem (student) & 0.993 & \textbf{69.6} & 0.952 & 43.3 & \textbf{0.925} & \textbf{20.6} \\
\bottomrule
\end{tabular}
}{\begin{tabular}{llcccccc}
\toprule
\rowcolor{fau}
\color{white}\textbf{Type} & \color{white}\textbf{Method} & \multicolumn{2}{c}{\color{white}\textbf{Copy detection}} & \multicolumn{2}{c}{\color{white}\textbf{Re-identification}} & \multicolumn{2}{c}{\color{white}\textbf{Degraded re-id}} \\
\cmidrule(lr){3-4}\cmidrule(lr){5-6}\cmidrule(lr){7-8}
 &  & AUC & R@5 & AUC & R@5 & AUC & R@5 \\
\midrule
\multirow{3}{*}{Pixel-wise} & MSE & 0.948 & 33.5 & 0.703 & 2.9 & 0.674 & 1.0 \\
 & NCC & 0.956 & 56.2 & 0.698 & 2.5 & 0.676 & 1.6 \\
 & SSIM & 0.942 & 40.8 & 0.663 & 4.2 & 0.642 & 1.9 \\
\midrule
\multirow{5}{*}{Pretrained} & CLIP & 0.856 & 4.7 & 0.695 & 1.2 & 0.634 & 0.0 \\
 & SwAV & 0.941 & 15.5 & 0.763 & 3.3 & 0.698 & 0.1 \\
 & DINOv2 & 0.904 & 17.0 & 0.702 & 1.4 & 0.644 & 0.0 \\
 & DINOv3 & 0.924 & 19.4 & 0.741 & 1.5 & 0.671 & 0.1 \\
 & VGG19 & 0.892 & 3.3 & 0.710 & 1.3 & 0.642 & 0.0 \\
\midrule
\multirow{2}{*}{Copy det.} & pHash & 0.945 & 56.9 & 0.620 & 0.9 & 0.606 & 0.6 \\
 & SSCD & \textbf{0.996} & 48.7 & 0.749 & 2.8 & 0.724 & 0.4 \\
\midrule
\multirow{8}{*}{Trained} & Siamese & 0.846 & 0.4 & 0.948 & 7.0 & 0.802 & 0.2 \\
 & Siamese + NT-Xent & 0.793 & 0.6 & 0.969 & 5.2 & 0.762 & 0.0 \\
 & Siamese + NT-Xent (unsup.) & 0.723 & 1.1 & 0.708 & 0.5 & 0.587 & 0.0 \\
 & Siamese (strong aug.) & 0.910 & 0.5 & 0.945 & 2.3 & 0.855 & 0.1 \\
 & Siamese + NT-Xent (strong aug.) & 0.908 & 5.3 & \textbf{0.970} & 5.0 & 0.871 & 0.4 \\
 & ArcFace & 0.871 & 20.1 & 0.956 & 33.1 & 0.772 & 1.4 \\
 & ArcFace (ConvNeXt-T) & 0.914 & 43.0 & 0.963 & 44.4 & 0.835 & 6.3 \\
 & ArcFace (strong aug.) & 0.871 & 21.1 & 0.874 & 10.8 & 0.736 & 2.2 \\
\midrule[1.6pt]
\rowcolor{black!12}Ours & LCMem (teacher) & 0.966 & 48.1 & 0.947 & \textbf{49.6} & 0.868 & 13.4 \\
\rowcolor{black!12} & LCMem (student) & 0.993 & \textbf{69.6} & 0.952 & 43.3 & \textbf{0.925} & \textbf{20.6} \\
\bottomrule
\end{tabular}
}}%
    \caption[Main results]{Main results on copy detection, re-identification and degraded re-identification. AUC: \AUCROC{} between positive and negative pairs; R@5: recall of the positive pairs at the threshold that flags 5\,\% of novel validation images (\cref{lcmem:sec:protocol}). Macro over the five datasets of \cref{lcmem:tab:dataset_overview}; copy detection and degraded re-identification are at corruption strength 4, averaged over the six transforms. Bold: best per column. Representative subset of the 26 baselines (all in \cref{lcmem:tab:main_full}). Row groups: pixel-wise, pretrained encoders, copy detection (hash, SSCD), trained on our data, ours. Both LCMem rows use the cosine of the embedding as pair score.}
    \label{lcmem:tab:main}
\end{table*}

\subsection{Copy Detection and Re-Identification}
\label{lcmem:sec:copydet}
\label{lcmem:sec:reid}
\label{lcmem:sec:corrupted_reid}
\noindent\textbf{Copy detection}
(\cref{lcmem:tab:main}, left columns) is led by SSCD, a method built for it (\AUCROC{} $0.996$), with the LCMem student on par ($0.993$); the pixel-wise metrics, pHash and the stronger pretrained encoders also separate copies well.
The models trained for re-identification are the weakest group, because nothing in their training rewards invariance to blur, noise or compression.
The operating point separates the methods further: the LCMem student recovers $69.6\,\%$ of the corrupted copies at R@5, against $56.9\,\%$ for pHash and $48.7\,\%$ for SSCD.
The gain of the student over the teacher ($48.1\,\%$) is the effect of the second stage alone, since the two models share every weight except the FiLM layers: the pixel-space augmentation and the latent noise during distillation (\cref{lcmem:sec:student}) make the embedding invariant to the corruptions without changing what it encodes.

\noindent\textbf{Re-identification:}
Re-identification is mostly a property depending on training on identity labels rather than which feature family was used (\cref{lcmem:tab:main}).
Every method trained on our identity labels reaches an \AUCROC{} above $0.94$, except ArcFace with strong augmentation ($0.874$), and no method without identity supervision passes $0.77$. Siamese + NT-Xent drops to $0.708$ when it is trained on the same data without the labels.
The \AUCROC{} hides the operating point: at R@5 the Siamese models recover at most $7.0\,\%$ of the positive pairs, because they separate positives from random negatives but not from the hardest negative of a large reference set, which is what the one-vs-all threshold demands.
The LCMem teacher has the highest recall of all methods ($49.6\,\%$), and the student keeps most of it ($43.3\,\%$, close to the best ArcFace model at $44.4\,\%$): the distillation on augmented and noised inputs costs little of the identity information of the teacher.

\noindent\textbf{Degraded re-identification}
combines both requirements: the positive pairs are two different images of the same identity, one of them corrupted.
This is the setting of a privacy filter in practice, where a generated image is neither a pixel copy of a training image nor guaranteed to be a clean one.
Here the two families fail in complementary ways (\cref{lcmem:tab:main}, right columns): the copy-detection methods lose the identity (R@5 below $2\,\%$), and the re-identification models lose the robustness (R@5 of at most $6.3\,\%$).
The LCMem student is first on both measures (\AUCROC{} $0.925$, R@5 $20.6\,\%$), with three times the recall of the best baseline, and the teacher is second in recall ($13.4\,\%$).
The three settings are therefore consistent: the teacher inherits re-identification from its identity training, and the student adds the invariance of a copy detector through distillation.
\Cref{lcmem:sec:operating_point,lcmem:sec:ova_deployment} show that this lead holds at every rejection budget and against the full training set.

\subsection{One-vs-All Operating Point}
\label{lcmem:sec:operating_point}

We sweep the oversampling cost $\rho$ from $1\,\%$ to $50\,\%$ and report, for every $\rho$, the recall of the positive test pairs at $\delta_\rho$, averaged over the five datasets.
\begin{figure*}[t]
    \centering
    \includegraphics[width=\linewidth]{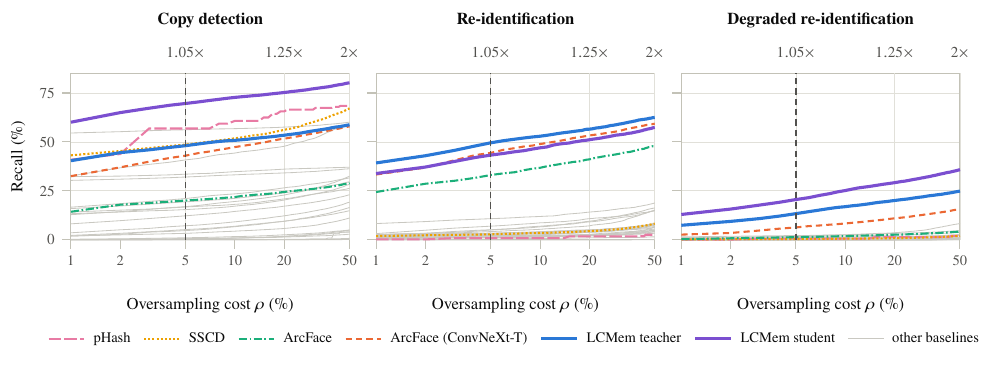}
    \caption{One-vs-all operating point. Recall of the positive test pairs against the oversampling cost $\rho$, the fraction of novel images the filter rejects (log scale; top axis: samples drawn per accepted image, $1/(1-\rho)$). Macro over the five datasets; copy detection and degraded re-identification at corruption strength 4, averaged over the six corruptions. Gray: the remaining 22 baselines. Dashed vertical line: $\rho = 5\,\%$, the operating point of all recall numbers in the tables.}
    \label{lcmem:fig:operating_point}
\end{figure*}

The recall of all methods grows roughly linearly in $\log \rho$ and does not saturate (\cref{lcmem:fig:operating_point}).
For the teacher on re-identification, raising $\rho$ from $5\,\%$ to $50\,\%$, which doubles the sampling cost, adds 13 points of recall ($49.6$ to $62.5\,\%$); copy detection and degraded re-identification grow at the same rate.
The ranking does not depend on the budget: the teacher is first for re-identification at every $\rho$, with the student five to six points below it, and the student is first for copy detection and degraded re-identification at every $\rho$.

\subsection{One-vs-All Deployment}
\label{lcmem:sec:ova_deployment}
We deploy LCMem as the privacy filter of a generative model that either reproduces a training image, re-generates a known identity, or synthesizes a novel identity.
The embeddings of all images of a split form an atlas, and every image of that split is used once as a query, clean and degraded by one of the six corruptions at strength 4 (fixed per image).
The query is scored by its highest similarity to the atlas.
A positive is a query that should be flagged: for copy detection, the degraded query whose own clean image is in the atlas; for re-identification and degraded re-identification, the clean or degraded query of an identity that has other images in the atlas (the query image itself is excluded).
The negative is the same query after its whole identity has been removed from the atlas.
We report the \AUCROC{} between the positive and best matches across all negatives in the split, averaged over the five datasets. We run this on the training split, the deployment case in which the atlas holds the training images of the generative model, and on the test split, which is disjoint at the identity level and shows the generalization to unseen identities.
Because the negative is a maximum over the atlas, it grows with the atlas size, so we also report the training split size-matched to the test split (mean over disjoint subsets of the training identities).
\emph{Inference speed.} For the student on CelebA, to check memorization of one batch of $8$ queries against the whole training dataset ($122{,}825$ training images) takes $10.5$\,ms on one A100 ($1.3$\,ms per image) and $0.48$\,GB of VRAM.

\emph{Results.} In \cref{lcmem:tab:ova_deployment}, the LCMem student is the best method for copy detection and degraded re-identification on every split.
For clean re-identification on the training split, the deployment case, the teacher is best ($.996$), ahead of the image-space ArcFace model and the student.
On unseen test identities, all trained models reach $.85$ to $.87$ for re-identification.
The exception is ISIC, where the test re-identification \AUCROC{} of every method is at most $.55$, while the other four datasets stay above $.87$ for the student: its images show a single skin lesion, so little of the patient's identity is visible, and patients that were not seen in training are hardly re-identified.
The copy-detection baselines pHash and SSCD do not re-identify, and the ArcFace models lose most of their performance once the query is degraded.

\emph{Strict filter.} 
At $\rho = 90\,\%$, i.e., ten times the sampling, the student catches $96.6$ to $99.7\,\%$ of the degraded copies on CelebA, MIMIC and NIH-CXR-LT and $92.3\,\%$ on CT-RATE, and $90.2$ to $92.7\,\%$ of the degraded re-identification queries on the same four datasets; clean re-identification is above $96\,\%$ on all five.
For strong privacy auditing, a tenfold increase in sampling is therefore enough to filter out about nine of ten copies of training images even after they have been degraded.
ISIC is the exception because its images carry little identity information (Full results \Cref{lcmem:sec:ova_budget}).

\begin{table}[t]
\centering
\renewcommand{\arraystretch}{1.1}
\small
\setlength{\tabcolsep}{3pt}
\resizebox{\linewidth}{!}{%
\begin{tabular}{lccccccccc}
\toprule
\rowcolor{fau}
\color{white} & \multicolumn{3}{c}{\color{white}\textbf{Train (full)}} & \multicolumn{3}{c}{\color{white}\textbf{Train (size-matched)}} & \multicolumn{3}{c}{\color{white}\textbf{Test}} \\
\rowcolor{fau}
\color{white}\textbf{Method} & \color{white}\textit{Copy} & \color{white}\textit{Re-ID} & \color{white}\textit{Deg.\ Re-ID} & \color{white}\textit{Copy} & \color{white}\textit{Re-ID} & \color{white}\textit{Deg.\ Re-ID} & \color{white}\textit{Copy} & \color{white}\textit{Re-ID} & \color{white}\textit{Deg.\ Re-ID} \\
\midrule
pHash & .737 & .195 & .181 & .755 & .219 & .203 & .757 & .228 & .212 \\
SSCD & .812 & .475 & .419 & .824 & .506 & .443 & .828 & .515 & .452 \\
ArcFace & .413 & .958 & .290 & .461 & .971 & .345 & .454 & .859 & .323 \\
ArcFace (ConvNeXt-T) & .621 & .988 & .502 & .658 & .991 & .555 & .623 & \textbf{.874} & .472 \\
\midrule
LCMem (teacher) & .635 & \textbf{.996} & .579 & .666 & \textbf{.998} & .623 & .645 & .863 & .536 \\
LCMem (student) & \textbf{.858} & .944 & \textbf{.767} & \textbf{.874} & .958 & \textbf{.801} & \textbf{.844} & .846 & \textbf{.729} \\
\bottomrule
\end{tabular}}
\caption{One-vs-all deployment. \AUCROC{} between the positive and the negative one-vs-all scores of all images of a split (strength 4), averaged over the five datasets.}
\label{lcmem:tab:ova_deployment}
\end{table}

\paragraph{Latent alignment and new autoencoders.}
We verified that the three training autoencoders are mapped to one joint embedding space: the same image in two autoencoders has a cosine similarity of $0.92$ to $0.99$ and is retrieved across spaces for $97$ to $100\,\%$ of the queries (\cref{lcmem:sec:latent_alignment}).
We also added the unseen FLUX.2 autoencoder by training only a new adapter ($0.02$M parameters, under one hour on precomputed latents), which reaches a re-identification \AUCROC{} of $0.964$, on par with the trained autoencoders ($0.963$ to $0.965$) after only 1000 steps (\cref{lcmem:tab:alignment_curve}; \cref{lcmem:sec:new_autoencoders}).

\begin{table}[t]
\centering
\renewcommand{\arraystretch}{1.1}
\small
\setlength{\tabcolsep}{4pt}
\resizebox{\linewidth}{!}{%
\begin{tabular}{rrccc}
\toprule
\rowcolor{fau}
\color{white}\textbf{Steps} & \color{white}\textbf{Images} & \color{white}\textbf{FLUX.2--FLUX.2} & \color{white}\textbf{SD2--FLUX.2} & \color{white}\textbf{Cos. similarity to target} \\
\midrule
0 (zero-shot) & 0 & .696 & .534 & .029 \\
30     & 1920    & .773 & .653 & .099 \\
100    & 6400    & .952 & .947 & .678 \\
300    & 19200   & .963 & .961 & .776 \\
1000   & 64000   & .964 & .962 & .806 \\
20000  & 1.28M   & .963 & .962 & .825 \\
20000, noisy latents & 1.28M & .962 & .960 & .813 \\
\midrule
\multicolumn{2}{l}{\textit{Reference: trained spaces}} & & & \\
\multicolumn{2}{l}{SD2 / SDXL / FLUX.1, same space} & .965 / .963 / .965 & -- & .816 / .814 / .817 \\
\bottomrule
\end{tabular}}
\caption{Adapter-only alignment of FLUX.2 to the frozen student. Validation \AUCROC{} (clean images, macro over the five datasets) with both images in FLUX.2 or with the reference in SD2 and the query in FLUX.2, and mean cosine similarity of the FLUX.2 embedding of an image to its atlas target. ``Noisy latents'': adapter trained with the latent noise of the student training. Bottom: the three trained autoencoders for reference.}
\label{lcmem:tab:alignment_curve}
\end{table}

\subsection{Applied Latent Memorization and Copy Detection}
\label{lcmem:sec:applied}
Next we want to show how LCMem can be used for deployment (training or sampling time).
The student is conditioned on the noise level, so it can be applied to the noisy latent $z_t$ of a sampler without decoding.
As generator we use an EDM2~\cite{karras_analyzing_2024} latent diffusion model in the FLUX.2 latent space, fine-tuned on the training images of CelebA and sampled with the deterministic Heun sampler with $T=32$ steps and noisy latents $z_t = z + \sigma_t\epsilon$ (\cref{lcmem:sec:generator}).
The student has never seen the FLUX.2 space in training, so every embedding in this section is computed by the student with the FLUX.2 adapter of \cref{lcmem:sec:new_autoencoders}: the adapter is the only trained part, it was fitted on noisy latents (last row of \cref{lcmem:tab:alignment_curve}), and all other weights, including the FiLM layers, are those of the student.
We test whether identity is already readable from $z_t$ during sampling, on 1000 CelebA training images and 1000 synthetic samples.

\begin{figure}[t]
    \centering
    \includegraphics[width=\linewidth]{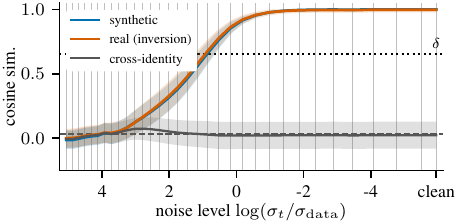}
    \caption{Student cosine similarity of the noisy latent $z_t$ to the clean reference over the sampling schedule (CelebA, 1000 images, mean $\pm$ std; grey lines: the 32 sampling steps). Dashed: mean cross-identity similarity; dotted: threshold $\delta$.}
    \label{lcmem:fig:applied}
\end{figure}

At every step $t$ we embed the state rescaled to the variance the student was trained with, $u_t = S_\psi(z_t / \sqrt{1 + \sigma_t^2}, \sigma_t)$, and score it by the cosine similarity to a reference embedding $u_{\mathrm{ref}}$ of a clean image.
For \emph{synthetic images}, $u_{\mathrm{ref}}$ is the embedding of the final sample, which measures from which step on the state is recognized as the image it becomes.
For \emph{real images}, $u_{\mathrm{ref}}$ is the embedding of the image $x$. We invert $z = \enc(x)$ to noise by integrating the sampling ODE in the reverse direction, sample again and score every intermediate state.
As a \emph{cross-identity} check, the same states are scored against the reference of a randomly drawn other image, which gives the level that identity-specific similarity has to exceed.
\Cref{lcmem:fig:applied} shows mean and standard deviation over the images for every step.

Of the 1000 inversions, one failed and was removed (reconstruction cosine below 0.5), and the curves of synthetic and real images are almost identical.
The threshold $\delta = 0.655$ is the one-vs-all threshold of the same FLUX.2-adapted student on CelebA (95th percentile of the best score of 2000 validation queries against all training images).
The mean similarity rises above $\delta$ at step 18 of 32 ($\sigma_t = 1.2$), and at step 20 ($\sigma_t = 0.6$) at least $99.5\,\%$ of the synthetic and of the real images are above it, while the cross-identity curve stays at a mean of 0.03 with at most $10^{-4}$ of its pairs above $\delta$ at any step.
Identity is therefore readable from the noisy state well before sampling ends. Memorized samples can be rejected after 20 of 32 steps, which saves the remaining steps and the decoding, and the same signal can steer latents away from memorized identities. %

\noindent\textbf{Limitations}
LCMem detects memorization but does not prevent it. At a strict operating point its recall is below 100\,\%, so some copies are missed; a filter built on it therefore reduces the risk of a leak but does not remove it (\cref{lcmem:tab:main}).
Re-identification of identities that were not seen in training fails on ISIC for every method (\cref{lcmem:sec:ova_deployment}).
The experiment on noisy sampler states (\cref{lcmem:sec:applied}) scores a state against the reference of its own image with a threshold fitted on clean images. 
LCMem does not address other risks such as membership inference attacks \cite{zhao2025does} and lacks interpretability and formal guarantees.

\section{Conclusion}
We framed memorization detection as one problem that spans copy detection, re-identification and degraded re-identification, and showed on five datasets that existing methods solve at most one part of it: copy detectors lose the identity, and re-identification models lose the robustness.
LCMem closes this gap with a single embedding network in the latent spaces of generative models, trained in two stages: a teacher that learns identities jointly across three autoencoders, and a noise-conditioned student distilled from it under augmentation and latent noise.
It has the highest recall in all three settings, works as a one-vs-all check against the full training set without decoding, extends to a new autoencoder by training one adapter, and reads the identity of a sample from the noisy latents of a diffusion sampler well before sampling ends.
Privacy can thus be enforced during sampling and, in the future, as a loss in the training of a generative model, in addition to filtering finished images.

\section*{LLM Usage Statement}
We used Anthropic's Claude models, through the Claude Code coding assistant, throughout this work. Affected parts: (i) the code of the experiments and evaluation pipeline (writing, refactoring and debugging); (ii) the plotting code of figures and tables; and (iii) the text of the paper, where the LLM drafted and edited passages and improved grammar and wording in all sections. The research ideas, the method, the experimental design and the interpretation of results are the authors'. The authors reviewed all generated code, results and text, verified all citations, and take full responsibility for the paper.

\section*{Acknowledgments}
HPC resources were provided by the Erlangen National High Performance Computing Center (NHR@FAU), under the NHR projects b180dc and y100dc. NHR is funded by federal and Bavarian state authorities, and NHR@FAU hardware is partially funded by the DFG, 440719683.
The authors received funding from the ERC-project MIA-NORMAL 101083647, 101326871, DFG 513220538, 512819079, and through the Hightech Agenda Bayern (HTA) of the Free State of Bavaria, Germany. This work was supported through the Bavarian AI Foundation Model Initiative, funded by the Bavarian State Ministry of Science and the Arts (StMWK).

{
    \small
    \bibliographystyle{ieeenat_fullname}
    \bibliography{references}
}

\clearpage
\appendix
\crefalias{section}{appendix}\crefalias{subsection}{subappendix}\crefalias{subsubsection}{subsubappendix}
\section{Appendix}
\providecommand{\lcmemtablabel}[1]{}\providecommand{\lcmemtablabelnostep}[1]{}

\subsection{Datasets}
\label{lcmem:sec:datasets}
\begin{table*}[t]
    \centering
    \renewcommand{\arraystretch}{1.2}
    \caption[Overview of the datasets used for evaluation]{Overview of the datasets used in our experiments. All datasets have identity labels, so copy detection, re-identification and degraded re-identification are evaluated on every dataset. The short name is used in all following tables and figures.}
    \resizebox{\textwidth}{!}{%
    \begin{tabular}{llccccccc}
    \toprule
    \rowcolor{fau}
    \color{white}\textbf{Dataset Name} & \color{white}\textbf{Short Name} & \color{white}\textbf{Medical} & \color{white}\textbf{Identity Label} & \color{white}\textbf{\# Images} & \color{white}\textbf{\# Identities} & \color{white}\textbf{Train / Val / Test} & \color{white}\textbf{Prediction Type} & \color{white}\textbf{\#(Labels/Classes)} \\
    \midrule
    CelebA \cite{liu2015deep} & CelebA & \xmark & \cmark & 202{,}599 & 10{,}177 & 122{,}825 / 19{,}760 / 60{,}014 & Labels & 40 facial attributes \\
    MIMIC-CXR \cite{johnson_mimic-cxr_2019} & MIMIC & \cmark & \cmark & 119{,}534 & 48{,}496 & 91{,}843 / 11{,}566 / 16{,}125 & Labels & 20 thoracic diseases classes \\
    NIH-CXR-LT \cite{nguyen_long-tailed_2022,wang2017chestx} & NIH & \cmark & \cmark & 88{,}637 & 29{,}460 & 68{,}058 / 300 / 20{,}279 & Classes &  20 thoracic diseases classes \\
    ISIC-2020 \cite{rotemberg_patient-centric_2021} & ISIC & \cmark & \cmark & 32{,}701 & 2{,}056 & 20{,}319 / 2{,}930 / 9{,}452 & Classes & Benign/Malignant classification \\
    CTRate \cite{hamamci_developing_2025} & CT-RATE & \cmark & \cmark & 25{,}049 & 2{,}652 & 14{,}852 / 2{,}538 / 7{,}659 & Labels & 15 pulmonary disease labels \\
    \bottomrule
    \end{tabular}}

    \label{lcmem:tab:dataset_overview}
\end{table*}

The selection of the datasets was conducted with the goal of combining medical and non-medical datasets, and we focused on datasets with identity labels present to enable training of the re-identification module.
An overview of all datasets used is shown in~\cref{lcmem:tab:dataset_overview}.
Image counts are unique images; the splits are disjoint at the image level and at the identity level.
ImageNet \cite{deng2009imagenet} (1.33M images, 1000 classes, no identity labels) is used as training data only.
All of it is used for the distillation of the initialisation (\cref{lcmem:sec:network}).
A class-balanced subset of $184{,}000$ images ($184$ per class, which matches the size of the balanced MIMIC-CXR training set) is used together with the five datasets in the training of the teacher, the student and the baselines that do not classify identities; there, every ImageNet image is its own identity.
ImageNet is not used for evaluation due to the lack of identity labels; all reported results are on the five datasets of \cref{lcmem:tab:dataset_overview}.
We train one joint model on all datasets, for LCMem and for every trained baseline.
Models trained on a single dataset perform on par (\cref{lcmem:sec:single_vs_multi}), so joint training gives the same performance with one training run instead of five.

\paragraph{Preprocessing of CT-RATE.}
The CTRate dataset consists of volumetric chest CT scans.
To enable training on two-dimensional images, we extract 2D slices that contain clinically described findings.
Using RexGrounding-CT annotations from Baharoon et al. 
\cite{baharoon_rexgroundingct_2025}, we identify and retain slices that either include or exclude abnormal findings.
For each finding instance, we extract all corresponding segmentation masks and include one representative image slice per instance in the final dataset.
To maintain balance, we also add a single slice without any labeled findings for each abnormal slice.
Note that such ``negative'' slices are not guaranteed to be healthy, as the labeling process in RexGrounding-CT is not exhaustive \cite{baharoon_rexgroundingct_2025}.
Each slice receives multi-label annotations according to the segmentation masks, allowing multiple co-occurring findings to be captured.

\subsection{Full Main Results}
\label{lcmem:sec:main_full}
\Cref{lcmem:tab:main_full} extends \cref{lcmem:tab:main} to all 26 baselines.
\begin{table*}[t]
    \centering\small
    \renewcommand{\arraystretch}{0.92}\setlength{\tabcolsep}{12pt}\footnotesize
    \resizebox{\textwidth}{!}{%
    \IfFileExists{figures/main_results_table.tex}{\begin{tabular}{llcccccc}
\toprule
\rowcolor{fau}
\color{white}\textbf{Type} & \color{white}\textbf{Method} & \multicolumn{2}{c}{\color{white}\textbf{Copy detection}} & \multicolumn{2}{c}{\color{white}\textbf{Re-identification}} & \multicolumn{2}{c}{\color{white}\textbf{Degraded re-id}} \\
\cmidrule(lr){3-4}\cmidrule(lr){5-6}\cmidrule(lr){7-8}
 &  & AUC & R@5 & AUC & R@5 & AUC & R@5 \\
\midrule
\multirow{4}{*}{Pixel-wise} & MSE & 0.948 & 33.5 & 0.703 & 2.9 & 0.674 & 1.0 \\
 & L1 & 0.953 & 32.1 & 0.712 & 3.3 & 0.683 & 1.0 \\
 & NCC & 0.956 & 56.2 & 0.698 & 2.5 & 0.676 & 1.6 \\
 & SSIM & 0.942 & 40.8 & 0.663 & 4.2 & 0.642 & 1.9 \\
\midrule
\multirow{11}{*}{Pretrained} & BYOL & 0.698 & 0.0 & 0.645 & 1.3 & 0.557 & 0.0 \\
 & CLIP & 0.856 & 4.7 & 0.695 & 1.2 & 0.634 & 0.0 \\
 & ConvNeXt & 0.889 & 7.2 & 0.710 & 2.0 & 0.658 & 0.0 \\
 & data2vec & 0.698 & 0.1 & 0.600 & 0.0 & 0.567 & 0.0 \\
 & Inception & 0.903 & 12.3 & 0.725 & 1.2 & 0.658 & 0.0 \\
 & MAE & 0.855 & 0.0 & 0.740 & 3.3 & 0.644 & 0.0 \\
 & SwAV & 0.941 & 15.5 & 0.763 & 3.3 & 0.698 & 0.1 \\
 & DINOv2 & 0.904 & 17.0 & 0.702 & 1.4 & 0.644 & 0.0 \\
 & DINOv3 & 0.924 & 19.4 & 0.741 & 1.5 & 0.671 & 0.1 \\
 & VGG19 & 0.892 & 3.3 & 0.710 & 1.3 & 0.642 & 0.0 \\
 & Random (untrained) & 0.824 & 16.8 & 0.663 & 2.0 & 0.605 & 0.3 \\
\midrule
\multirow{2}{*}{Copy det.} & pHash & 0.945 & 56.9 & 0.620 & 0.9 & 0.606 & 0.6 \\
 & SSCD & \textbf{0.996} & 48.7 & 0.749 & 2.8 & 0.724 & 0.4 \\
\midrule
\multirow{8}{*}{Trained} & Siamese & 0.846 & 0.4 & 0.948 & 7.0 & 0.802 & 0.2 \\
 & Siamese + NT-Xent & 0.793 & 0.6 & 0.969 & 5.2 & 0.762 & 0.0 \\
 & Siamese + NT-Xent (unsup.) & 0.723 & 1.1 & 0.708 & 0.5 & 0.587 & 0.0 \\
 & Siamese (strong aug.) & 0.910 & 0.5 & 0.945 & 2.3 & 0.855 & 0.1 \\
 & Siamese + NT-Xent (strong aug.) & 0.908 & 5.3 & \textbf{0.970} & 5.0 & 0.871 & 0.4 \\
 & ArcFace & 0.871 & 20.1 & 0.956 & 33.1 & 0.772 & 1.4 \\
 & ArcFace (ConvNeXt-T) & 0.914 & 43.0 & 0.963 & 44.4 & 0.835 & 6.3 \\
 & ArcFace (strong aug.) & 0.871 & 21.1 & 0.874 & 10.8 & 0.736 & 2.2 \\
\midrule[1.6pt]
\rowcolor{black!12}Ours & LCMem (teacher) & 0.966 & 48.1 & 0.947 & \textbf{49.6} & 0.868 & 13.4 \\
\rowcolor{black!12} & LCMem (student) & 0.993 & \textbf{69.6} & 0.952 & 43.3 & \textbf{0.925} & \textbf{20.6} \\
\bottomrule
\end{tabular}
}{}}
    \caption[Full main results]{Full version of \cref{lcmem:tab:main} with all baselines. Same metrics and protocol. Row groups: pixel-wise, pretrained encoders (incl.\ VGG19 and the untrained Random baseline), copy detection (hash, SSCD), trained on our data, ours.}
    \label{lcmem:tab:main_full}
\end{table*}

\subsection{Results per Dataset}
\label{lcmem:sec:per_dataset}
\Cref{lcmem:tab:per_dataset_main,lcmem:tab:per_dataset_ova} break down the macro results of \cref{lcmem:tab:main} and of the one-vs-all deployment table into the five datasets, for the baselines of the deployment table, two trained baselines with strong augmentation, and LCMem.
The macro over the five datasets of each column reproduces the corresponding main-paper table.
Copy detection and degraded re-identification are at corruption strength 4 (mean over the six corruptions).

\begin{table*}[t]
    \lcmemtablabel{6a}
    \centering\small
    \renewcommand{\arraystretch}{1.0}\setlength{\tabcolsep}{10pt}\footnotesize
    \IfFileExists{figures/supp_dataset_main.tex}{\begin{tabular}{lcccccc}
\toprule
\rowcolor{fau}
\color{white}\textbf{Method} & \multicolumn{2}{c}{\color{white}\textbf{Copy detection}} & \multicolumn{2}{c}{\color{white}\textbf{Re-identification}} & \multicolumn{2}{c}{\color{white}\textbf{Degraded re-id}} \\
\cmidrule(lr){2-3}\cmidrule(lr){4-5}\cmidrule(lr){6-7}
 & AUC & R@5 & AUC & R@5 & AUC & R@5 \\
\midrule
\multicolumn{7}{l}{\textit{CelebA}} \\
pHash & 0.954 & 67.1 & 0.581 & 0.0 & 0.577 & 0.0 \\
SSCD & \textbf{1.000} & \textbf{97.3} & 0.757 & 1.8 & 0.747 & 1.0 \\
ArcFace & 0.915 & 49.0 & 0.978 & 25.8 & 0.864 & 4.9 \\
ArcFace (ConvNeXt-T) & 0.929 & 66.8 & \textbf{0.981} & \textbf{35.3} & 0.890 & 11.3 \\
Siamese + NT-Xent (strong aug.) & 0.977 & 12.7 & 0.964 & 1.1 & 0.929 & 0.2 \\
ArcFace (strong aug.) & 0.912 & 26.7 & 0.778 & 0.7 & 0.691 & 0.2 \\
\rowcolor{black!12}LCMem (teacher) & 0.989 & 62.6 & 0.978 & 31.4 & 0.923 & \textbf{11.6} \\
\rowcolor{black!12}LCMem (student) & \textbf{1.000} & 88.7 & 0.978 & 19.9 & \textbf{0.968} & 9.7 \\
\midrule
\multicolumn{7}{l}{\textit{MIMIC-CXR}} \\
pHash & 0.928 & 46.8 & 0.578 & 0.1 & 0.562 & 0.1 \\
SSCD & 0.998 & 28.4 & 0.783 & 1.2 & 0.750 & 0.1 \\
ArcFace & 0.892 & 25.3 & 0.987 & 32.6 & 0.792 & 0.2 \\
ArcFace (ConvNeXt-T) & 0.931 & 65.6 & 0.992 & 56.8 & 0.874 & 6.5 \\
Siamese + NT-Xent (strong aug.) & 0.925 & 1.4 & 0.993 & 1.6 & 0.906 & 0.1 \\
ArcFace (strong aug.) & 0.922 & 27.0 & 0.881 & 1.0 & 0.749 & 0.2 \\
\rowcolor{black!12}LCMem (teacher) & 0.986 & 66.5 & \textbf{0.997} & \textbf{73.1} & 0.940 & 13.1 \\
\rowcolor{black!12}LCMem (student) & \textbf{1.000} & \textbf{95.6} & \textbf{0.997} & 65.7 & \textbf{0.991} & \textbf{25.3} \\
\midrule
\multicolumn{7}{l}{\textit{NIH-CXR-LT}} \\
pHash & 0.923 & 48.3 & 0.557 & 0.0 & 0.549 & 0.0 \\
SSCD & 0.989 & 33.5 & 0.738 & 3.9 & 0.700 & 0.2 \\
ArcFace & 0.877 & 24.1 & 0.988 & 42.9 & 0.796 & 1.7 \\
ArcFace (ConvNeXt-T) & 0.916 & 59.2 & \textbf{0.996} & 64.5 & 0.869 & 10.0 \\
Siamese + NT-Xent (strong aug.) & 0.901 & 1.7 & 0.992 & 4.4 & 0.878 & 0.1 \\
ArcFace (strong aug.) & 0.889 & 23.5 & 0.877 & 1.4 & 0.730 & 0.2 \\
\rowcolor{black!12}LCMem (teacher) & 0.961 & 58.0 & \textbf{0.996} & \textbf{76.1} & 0.914 & 19.6 \\
\rowcolor{black!12}LCMem (student) & \textbf{0.999} & \textbf{90.9} & \textbf{0.996} & 67.8 & \textbf{0.986} & \textbf{32.6} \\
\midrule
\multicolumn{7}{l}{\textit{ISIC-2020}} \\
pHash & 0.959 & \textbf{50.8} & 0.656 & 0.0 & 0.634 & 0.0 \\
SSCD & \textbf{0.991} & 38.3 & 0.768 & 0.0 & 0.740 & 0.0 \\
ArcFace & 0.783 & 1.1 & 0.886 & \textbf{2.7} & 0.631 & 0.0 \\
ArcFace (ConvNeXt-T) & 0.853 & 2.6 & 0.882 & 2.4 & 0.678 & 0.0 \\
Siamese + NT-Xent (strong aug.) & 0.908 & 0.7 & \textbf{0.954} & 0.1 & \textbf{0.860} & 0.0 \\
ArcFace (strong aug.) & 0.819 & 0.1 & 0.879 & 0.1 & 0.763 & 0.0 \\
\rowcolor{black!12}LCMem (teacher) & 0.913 & 1.5 & 0.848 & 2.2 & 0.694 & 0.1 \\
\rowcolor{black!12}LCMem (student) & 0.969 & 2.6 & 0.854 & 1.4 & 0.766 & \textbf{0.2} \\
\midrule
\multicolumn{7}{l}{\textit{CT-RATE}} \\
pHash & 0.959 & \textbf{71.3} & 0.726 & 4.5 & 0.706 & 3.2 \\
SSCD & \textbf{1.000} & 45.8 & 0.699 & 7.3 & 0.684 & 0.9 \\
ArcFace & 0.887 & 0.9 & 0.941 & 61.5 & 0.777 & 0.1 \\
ArcFace (ConvNeXt-T) & 0.939 & 20.7 & \textbf{0.966} & 63.0 & 0.863 & 3.8 \\
Siamese + NT-Xent (strong aug.) & 0.829 & 9.8 & 0.946 & 18.0 & 0.782 & 1.5 \\
ArcFace (strong aug.) & 0.811 & 28.2 & 0.958 & 50.9 & 0.749 & 10.4 \\
\rowcolor{black!12}LCMem (teacher) & 0.981 & 51.8 & 0.917 & \textbf{65.0} & 0.871 & 22.6 \\
\rowcolor{black!12}LCMem (student) & 0.999 & 70.3 & 0.936 & 61.6 & \textbf{0.913} & \textbf{35.2} \\
\bottomrule
\end{tabular}
}{}
    \caption{Results per dataset (TEST). Same metrics, protocol and thresholds as \cref{lcmem:tab:main}: AUC, and recall (\%) at the threshold that flags 5\,\% of novel validation images (R@5). Best value per dataset and column in bold.}
    \label{lcmem:tab:per_dataset_main}
\end{table*}

\begin{table*}[t]
    \lcmemtablabelnostep{6b}
    \centering\small
    \renewcommand{\arraystretch}{1.0}\setlength{\tabcolsep}{8pt}\footnotesize
    \IfFileExists{figures/supp_dataset_ova.tex}{\begin{tabular}{lccccccccc}
\toprule
\rowcolor{fau}
\color{white}\textbf{Method} & \multicolumn{3}{c}{\color{white}\textbf{Train (full)}} & \multicolumn{3}{c}{\color{white}\textbf{Train (size-matched)}} & \multicolumn{3}{c}{\color{white}\textbf{Test}} \\
\cmidrule(lr){2-4}\cmidrule(lr){5-7}\cmidrule(lr){8-10}
 & Copy & Re-ID & Deg. & Copy & Re-ID & Deg. & Copy & Re-ID & Deg. \\
\midrule
\multicolumn{10}{l}{\textit{CelebA}} \\
pHash & 0.794 & 0.078 & 0.071 & 0.801 & 0.090 & 0.082 & 0.802 & 0.092 & 0.083 \\
SSCD & \textbf{0.996} & 0.370 & 0.347 & \textbf{0.998} & 0.397 & 0.374 & \textbf{0.998} & 0.395 & 0.374 \\
ArcFace & 0.673 & 0.927 & 0.546 & 0.687 & 0.943 & 0.567 & 0.679 & 0.887 & 0.531 \\
ArcFace (ConvNeXt-T) & 0.788 & 0.965 & 0.699 & 0.795 & 0.975 & 0.716 & 0.787 & \textbf{0.912} & 0.642 \\
\rowcolor{black!12}LCMem (teacher) & 0.701 & \textbf{0.991} & 0.678 & 0.712 & \textbf{0.995} & 0.692 & 0.698 & 0.898 & 0.597 \\
\rowcolor{black!12}LCMem (student) & 0.968 & 0.927 & \textbf{0.811} & 0.975 & 0.945 & \textbf{0.841} & 0.973 & 0.876 & \textbf{0.759} \\
\midrule
\multicolumn{10}{l}{\textit{MIMIC-CXR}} \\
pHash & 0.666 & 0.062 & 0.070 & 0.705 & 0.094 & 0.100 & 0.709 & 0.117 & 0.123 \\
SSCD & 0.801 & 0.443 & 0.386 & 0.821 & 0.493 & 0.426 & 0.831 & 0.521 & 0.448 \\
ArcFace & 0.431 & 0.904 & 0.129 & 0.532 & 0.939 & 0.247 & 0.507 & 0.912 & 0.242 \\
ArcFace (ConvNeXt-T) & 0.702 & 0.979 & 0.427 & 0.760 & 0.987 & 0.512 & 0.702 & 0.969 & 0.461 \\
\rowcolor{black!12}LCMem (teacher) & 0.719 & \textbf{0.992} & 0.561 & 0.770 & \textbf{0.995} & 0.659 & 0.754 & \textbf{0.986} & 0.603 \\
\rowcolor{black!12}LCMem (student) & \textbf{0.991} & 0.972 & \textbf{0.845} & \textbf{0.995} & 0.981 & \textbf{0.895} & \textbf{0.994} & 0.983 & \textbf{0.894} \\
\midrule
\multicolumn{10}{l}{\textit{NIH-CXR-LT}} \\
pHash & 0.687 & 0.080 & 0.110 & 0.714 & 0.111 & 0.141 & 0.720 & 0.128 & 0.159 \\
SSCD & 0.711 & 0.506 & 0.431 & 0.728 & 0.545 & 0.458 & 0.740 & 0.562 & 0.472 \\
ArcFace & 0.451 & 0.972 & 0.313 & 0.487 & 0.981 & 0.353 & 0.483 & 0.957 & 0.343 \\
ArcFace (ConvNeXt-T) & 0.661 & 0.994 & 0.485 & 0.727 & 0.996 & 0.583 & 0.688 & 0.984 & 0.514 \\
\rowcolor{black!12}LCMem (teacher) & 0.674 & \textbf{0.998} & 0.590 & 0.714 & \textbf{0.998} & 0.645 & 0.712 & \textbf{0.993} & 0.626 \\
\rowcolor{black!12}LCMem (student) & \textbf{0.956} & 0.987 & \textbf{0.881} & \textbf{0.961} & 0.991 & \textbf{0.904} & \textbf{0.964} & 0.989 & \textbf{0.903} \\
\midrule
\multicolumn{10}{l}{\textit{ISIC-2020}} \\
pHash & 0.725 & 0.118 & 0.107 & 0.736 & 0.143 & 0.129 & 0.732 & 0.135 & 0.123 \\
SSCD & \textbf{0.726} & 0.330 & 0.351 & \textbf{0.741} & 0.367 & 0.372 & \textbf{0.741} & 0.358 & \textbf{0.368} \\
ArcFace & 0.243 & 0.989 & 0.197 & 0.272 & 0.992 & 0.235 & 0.270 & \textbf{0.546} & 0.171 \\
ArcFace (ConvNeXt-T) & 0.309 & \textbf{1.000} & 0.280 & 0.339 & \textbf{1.000} & 0.318 & 0.301 & 0.510 & 0.142 \\
\rowcolor{black!12}LCMem (teacher) & 0.406 & \textbf{1.000} & 0.387 & 0.439 & \textbf{1.000} & 0.424 & 0.382 & 0.442 & 0.172 \\
\rowcolor{black!12}LCMem (student) & 0.495 & 0.833 & \textbf{0.425} & 0.543 & 0.873 & \textbf{0.475} & 0.419 & 0.385 & 0.225 \\
\midrule
\multicolumn{10}{l}{\textit{CT-RATE}} \\
pHash & 0.812 & 0.639 & 0.546 & 0.820 & 0.656 & 0.560 & 0.819 & 0.669 & 0.570 \\
SSCD & 0.826 & 0.727 & 0.579 & 0.833 & 0.728 & 0.585 & 0.831 & 0.740 & 0.598 \\
ArcFace & 0.266 & \textbf{1.000} & 0.263 & 0.325 & \textbf{1.000} & 0.323 & 0.330 & 0.995 & 0.328 \\
ArcFace (ConvNeXt-T) & 0.644 & \textbf{1.000} & 0.617 & 0.668 & \textbf{1.000} & 0.642 & 0.635 & 0.995 & 0.599 \\
\rowcolor{black!12}LCMem (teacher) & 0.678 & \textbf{1.000} & 0.679 & 0.693 & \textbf{1.000} & 0.696 & 0.681 & \textbf{0.997} & 0.680 \\
\rowcolor{black!12}LCMem (student) & \textbf{0.880} & 0.999 & \textbf{0.875} & \textbf{0.894} & 0.999 & \textbf{0.890} & \textbf{0.870} & 0.996 & \textbf{0.864} \\
\bottomrule
\end{tabular}
}{}
    \caption{One-vs-all deployment per dataset. \AUCROC{} between the positive and the negative one-vs-all scores of all images of a split (strength 4), as in \cref{lcmem:tab:ova_deployment}. Train (full): the full training split as atlas; size-matched: mean over disjoint identity subsets of the size of the test split (equal to ``full'' for splits not larger than the test split); Test: the test split. Copy: copy detection, Re-ID: re-identification, Deg.: degraded re-identification. Best value per dataset and column in bold.}
    \label{lcmem:tab:per_dataset_ova}
\end{table*}

\subsection{Evaluation Protocol Details}
\label{lcmem:sec:protocol_details}
\begin{figure*}[t]
    \centering
    \includegraphics[width=\linewidth]{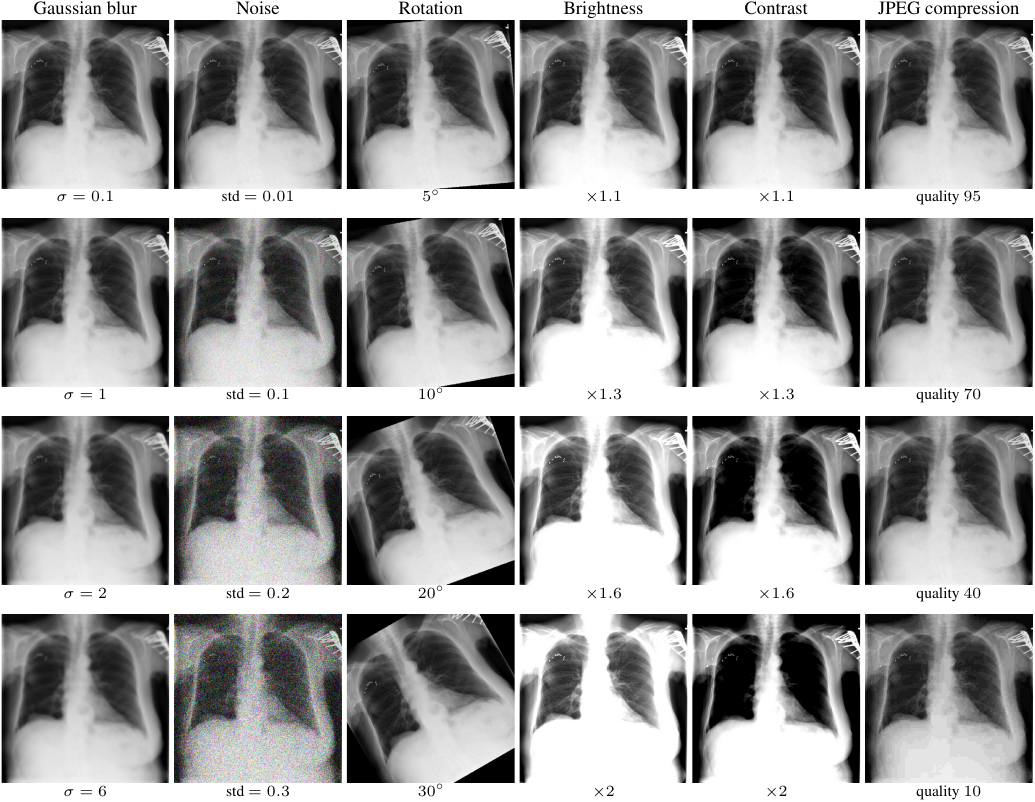}
    \caption{The six corruptions $a \in \augset$ at strengths 1 to 4 (top to bottom), applied to an NIH-CXR-LT image after resizing to $512 \times 512$. Each panel is labeled with the parameter of the table above; noise is Gaussian with the given standard deviation, rotation leaves the corners black.}
    \label{lcmem:fig:corruptions}
\end{figure*}

Strengths of the corruptions $a \in \augset$, applied to the second image of a pair after resizing to $512 \times 512$ and scaling to $[0,1]$:
\begin{center}
\renewcommand{\arraystretch}{1.2}
\footnotesize
\setlength{\tabcolsep}{3pt}
\resizebox{\linewidth}{!}{%
\begin{tabular}{lcccc}
\toprule
\rowcolor{fau}
\color{white}\textbf{Transform} & \color{white}\textbf{1} & \color{white}\textbf{2} & \color{white}\textbf{3} & \color{white}\textbf{4} \\
\midrule
Gaussian blur (kernel 9, std) & 0.1 & 1 & 2 & 6 \\
Gaussian noise (std, clamped to $[0,1]$) & 0.01 & 0.1 & 0.2 & 0.3 \\
Rotation (degrees) & 5 & 10 & 20 & 30 \\
Brightness factor & 1.1 & 1.3 & 1.6 & 2 \\
Contrast factor & 1.1 & 1.3 & 1.6 & 2 \\
JPEG compression (quality $= 100 - $ strength) & 5 & 30 & 60 & 90 \\
\bottomrule
\end{tabular}}
\end{center}
Positive pairs of the three settings share their anchors $x$, and the pairs are identical for all methods, transforms and strengths.
Pairs are always built from unique images; oversampled copies used for training never enter pairs or thresholds.
\Cref{lcmem:fig:corruptions} shows the six corruptions at all four strengths.

\subsection{Single vs.\ Multi-Dataset Evaluation}
\label{lcmem:sec:single_vs_multi}
Our models are trained on all datasets at once.
We investigate (a) whether training on multiple datasets is beneficial for learning features that generalize across datasets, and (b) whether training on all datasets at once performs roughly on par with training one model per dataset, which would make experimentation cheaper.
We use the two strongest baselines, ArcFace and Siamese + NT-Xent with strong augmentation, and train five models for each of them, one per dataset (CelebA, MIMIC, NIH-CXR-LT, ISIC, CT-RATE), with the same hyperparameters, augmentation, and early stopping on validation \AUCROC{} as the model trained on all datasets.
Each single-dataset model is evaluated on the test split of its own dataset with the protocol of the main paper; the macro average is taken over the five models.
\Cref{lcmem:tab:single_vs_multi} reports the \AUCROC{} for copy detection, re-identification, and degraded re-identification, the first and the last at strength 4 and averaged over the six corruptions.

For re-identification, the single-dataset models and the model trained on all datasets perform almost identically (macro \AUCROC{} 0.964 vs.\ 0.956 for ArcFace, 0.977 vs.\ 0.970 for Siamese + NT-Xent).
Under corruption the macro averages are close as well: for copy detection the model trained on all datasets is slightly ahead (0.871 vs.\ 0.846 for ArcFace, 0.908 vs.\ 0.899 for Siamese + NT-Xent), and for degraded re-identification the two variants are within 0.003 (0.772 vs.\ 0.769 and 0.871 vs.\ 0.873).
Per dataset the picture is mixed.
The single-dataset models are better on CelebA, and for Siamese + NT-Xent also on the two chest X-ray datasets (copy detection on NIH-CXR-LT: 0.983 vs.\ 0.901), but clearly worse on CT-RATE, the smallest dataset (Siamese + NT-Xent: 0.721 vs.\ 0.829 for copy detection and 0.687 vs.\ 0.782 for degraded re-identification).
Training on all datasets at once therefore gives the same performance on average at the cost of one training run instead of five.
All numbers come from a single training run per model.
The Siamese + NT-Xent model trained on all datasets additionally sees the ImageNet subset, which the single-dataset models do not; ArcFace needs identity labels and is trained without ImageNet in both cases.

\begin{table}[ht]
\lcmemtablabelnostep{6c}
\centering
\renewcommand{\arraystretch}{1.1}
\scriptsize
\setlength{\tabcolsep}{2pt}
\resizebox{\linewidth}{!}{%
\begin{tabular}{llcccccc}
\toprule
\rowcolor{fau}
\color{white}\textbf{Setting} & \color{white}\textbf{Method} & \color{white}\textbf{CA} & \color{white}\textbf{MI} & \color{white}\textbf{CX} & \color{white}\textbf{IS} & \color{white}\textbf{CT} & \color{white}\textbf{Macro} \\
\midrule
Copy & ArcFace, all datasets & .915 & .892 & .877 & .783 & .887 & .871 \\
    & ArcFace, single & .923 & .842 & .855 & .818 & .791 & .846 \\
    & Siamese + NT-Xent, all datasets & .977 & .925 & .901 & .908 & .829 & .908 \\
    & Siamese + NT-Xent, single & .996 & .947 & .983 & .848 & .721 & .899 \\
\midrule
Re-ID & ArcFace, all datasets & .978 & .987 & .988 & .886 & .941 & .956 \\
    & ArcFace, single & .984 & .998 & .997 & .893 & .949 & .964 \\
    & Siamese + NT-Xent, all datasets & .964 & .993 & .992 & .954 & .946 & .970 \\
    & Siamese + NT-Xent, single & .985 & .998 & .999 & .958 & .947 & .977 \\
\midrule
Deg.\ Re-ID & ArcFace, all datasets & .864 & .792 & .796 & .631 & .777 & .772 \\
    & ArcFace, single & .897 & .761 & .816 & .662 & .708 & .769 \\
    & Siamese + NT-Xent, all datasets & .929 & .906 & .878 & .860 & .782 & .871 \\
    & Siamese + NT-Xent, single & .973 & .939 & .975 & .793 & .687 & .873 \\
\bottomrule
\end{tabular}}
\caption{Single vs.\ multi-dataset training. \AUCROC{} of ArcFace and Siamese + NT-Xent (strong augmentation) trained on all datasets at once (``all datasets'') or with one model per dataset (``single''; macro over the five models). Copy: copy detection, Re-ID: re-identification, Deg.\ Re-ID: degraded re-identification; copy detection and degraded re-identification at corruption strength 4, averaged over the six corruptions. CA: CelebA, MI: MIMIC, CX: NIH-CXR-LT, IS: ISIC, CT: CT-RATE.}
\label{lcmem:tab:single_vs_multi}
\end{table}

\subsection{Ablations}
\label{lcmem:sec:ablations}
\Cref{lcmem:tab:ablation} ablates the teacher: the input space, the initialisation, the stride of the stem, the loss terms and the backbone.
All rows are evaluated on the test split with the cosine score; model selection used the validation split only.

\begin{table*}[t]
\centering
\renewcommand{\arraystretch}{1.1}
\small
\setlength{\tabcolsep}{5pt}
\resizebox{\linewidth}{!}{%
\begin{tabular}{llccccccc}
\toprule
\rowcolor{fau}
\color{white}\textbf{Group} & \color{white}\textbf{Variant} & \color{white}\textbf{Init} & \color{white}\textbf{BCE} & \color{white}\textbf{NCE} & \color{white}\textbf{Atlas} & \color{white}\textbf{Copy} & \color{white}\textbf{Re-ID} & \color{white}\textbf{Deg.\ Re-ID} \\
\midrule
Space (ArcFace only) & Image space (ConvNeXt-T, $512^2$ px, same recipe as latents) & IN & \xmark & \xmark & \xmark & .902 & .951 & .817 \\
      & Image space (ConvNeXt-T, $512^2$ px, baseline recipe) & IN & \xmark & \xmark & \xmark & .914 & .963 & .835 \\
\midrule
Init (stride 1) & ImageNet-pretrained & IN & \xmark & \xmark & \xmark & .856 & .909 & .735 \\
      & Distilled (ConvNeXt-L) & dist. & \xmark & \xmark & \xmark & .951 & .937 & .847 \\
\midrule
Stem (ArcFace only) & Stride 4 ($16^2$ grid) & IN & \xmark & \xmark & \xmark & .889 & .858 & .712 \\
      & Stride 2 ($32^2$ grid) & IN & \xmark & \xmark & \xmark & .873 & .880 & .716 \\
      & Stride 1 ($64^2$ grid) & IN & \xmark & \xmark & \xmark & .856 & .909 & .735 \\
\midrule
Loss (distilled, stride 1) & ArcFace only        & dist. & \xmark & \xmark & \xmark & .951 & .937 & .847 \\
      & + BCE                 & dist. & \cmark & \xmark & \xmark & .952 & .945 & .851 \\
      & + NCE                 & dist. & \xmark & \cmark & \xmark & .961 & .943 & .859 \\
      & + Atlas               & dist. & \xmark & \xmark & \cmark & .965 & .945 & .861 \\
      & + BCE + NCE           & dist. & \cmark & \cmark & \xmark & .961 & .946 & .859 \\
      & + BCE + Atlas         & dist. & \cmark & \xmark & \cmark & .954 & .939 & .851 \\
      & + NCE + Atlas         & dist. & \xmark & \cmark & \cmark & .956 & .946 & .856 \\
      & + BCE + NCE + Atlas   & dist. & \cmark & \cmark & \cmark & .961 & .944 & .861 \\
\midrule
\textbf{Full} & + drop path 0.3, weight decay 0.05 (ConvNeXt-T) & dist. & \xmark & \cmark & \cmark & \textbf{.966} & \textbf{.947} & \textbf{.868} \\
Full, large & DINOv2 ViT-B/14, 512-d embedding & dist. & \xmark & \cmark & \cmark & .976 & .956 & .894 \\
      & DINOv2 ViT-B/14, 768-d embedding & dist. & \xmark & \cmark & \cmark & .977 & .956 & .894 \\
\bottomrule
\end{tabular}}
\caption{Ablations (\AUCROC{}, macro over the five datasets, TEST). Copy: copy detection, Re-ID: re-identification, Deg.\ Re-ID: degraded re-identification; copy and degraded at corruption strength 4. All rows use ArcFace; BCE, NCE and Atlas mark the added losses. ``IN'': ImageNet-pretrained body, ``dist.'': body distilled from ConvNeXt-L on ImageNet. The loss rows share the distilled init and stem stride 1 and do not use the regularisation of the last row. The stem rows use ArcFace only with the ImageNet-pretrained body. The first image-space row uses the recipe of the latent rows (ArcFace only, per-dataset heads, batch 256 pairs, no augmentation, same pairs and splits) on RGB pixels with the original ConvNeXt stem (no adapter); the second is the ArcFace baseline of \cref{lcmem:tab:main} (batch 128, one head, augmentation). ``Full, large'' replaces the ConvNeXt-T (28\,M parameters) of the full model by a DINOv2 ViT-B/14 (86\,M) with patch size 4 on the adapter output, distilled from the pixel DINOv2 ViT-B/14 on ImageNet; the embedding is the CLS token, projected to 512 dimensions or used directly (768). They use a lower learning rate for the body ($2\cdot10^{-5}$, 1000 warm-up steps), because training diverged with the learning rate of the other rows. Bold marks the model used in all other experiments. One training run per row.}
\label{lcmem:tab:ablation}
\end{table*}

\paragraph{Input space.}
With the same recipe, the image-space teacher matches the latent teacher on clean re-identification ($.951$ vs.\ $.937$ \AUCROC{}) but is behind on copy detection ($.902$ vs.\ $.951$) and degraded re-identification ($.817$ vs.\ $.847$).
The latent space alone does not explain this: with the ImageNet-pretrained body, the latent model is below the image-space model on both ($.856$ and $.735$).
The robustness appears only together with the distilled body, which is trained on the latents of the three autoencoders and the adapters in front of it.
We do not separate the contribution of the frozen autoencoder encoders from that of the adapters; this would need a latent model without adapter and an image-space model with one.

\paragraph{Backbone.}
We keep the ConvNeXt-T teacher for all further experiments: it has the same backbone as the strongest trained baseline (ArcFace ConvNeXt-T) and is not larger than any other trained model.
In \cref{lcmem:tab:ablation}, we show that our training benefits from a larger backbone: a DINOv2 ViT-B/14 with three times the parameters improves all three settings ($.976$ / $.956$ / $.894$ vs.\ $.966$ / $.947$ / $.868$ \AUCROC{}).
The size of the embedding has no effect ($512$ vs.\ $768$ dimensions).
The gain is not uniform: copy detection on CT-RATE is worse than with the ConvNeXt-T ($.970$ vs.\ $.981$ \AUCROC{} at strength 4).

\subsection{Latent Alignment}
\label{lcmem:sec:latent_alignment}
We first test whether the three training autoencoders (SD2, SDXL, FLUX.1) are mapped to one joint embedding space.
For 1000 validation images per dataset, we embed each image through each of the three latent spaces with the student (clean conditioning) and compare the embeddings with the cosine similarity, averaged over the three space pairs (\Cref{lcmem:tab:cross_space}).
The same image in two different autoencoders is embedded almost identically (cosine $0.92$ to $0.99$), and retrieving it among the 1000 images of the other space succeeds for $97$ to $100\,\%$ of the queries.
Other images of the same identity in another space are as close as in the same space (e.g.\ $0.51$ vs.\ $0.51$ on CelebA), and different identities are far apart ($\leq 0.05$ on all re-identification datasets).
The autoencoder therefore carries almost no identity-irrelevant information into the embedding, which is what allows an atlas computed in one space to be queried from another.
ISIC has the lowest agreement ($0.92$); it is also the dataset with the lowest re-identification \AUCROC{}.

\begin{table}[t]
\centering
\renewcommand{\arraystretch}{1.1}
\small
\setlength{\tabcolsep}{3pt}
\resizebox{\linewidth}{!}{%
\begin{tabular}{lccccc}
\toprule
\rowcolor{fau}
\color{white}\textbf{Dataset} & \color{white}\textbf{Same image} & \color{white}\textbf{Top-1} & \color{white}\textbf{Same id.} & \color{white}\textbf{Same id.} & \color{white}\textbf{Diff.\ id.} \\
\rowcolor{fau}
\color{white} & \color{white}\textit{other space} & \color{white}\textit{other space} & \color{white}\textit{other space} & \color{white}\textit{same space} & \color{white}\textit{other space} \\
\midrule
CelebA      & .977 & 1.000 & .510 & .513 & .034 \\
MIMIC       & .972 & 1.000 & .596 & .605 & .041 \\
NIH-CXR-LT  & .972 & 1.000 & .642 & .652 & .049 \\
ISIC        & .920 & .986  & .321 & .330 & .036 \\
CT-RATE     & .989 & .972  & .712 & .717 & .031 \\
\bottomrule
\end{tabular}}
\caption{Agreement of the embeddings across autoencoders (cosine similarity, mean over the pairs SD2--SDXL, SD2--FLUX.1, SDXL--FLUX.1; 1000 validation images per dataset). ``Same image'': the same image encoded by two autoencoders; ``Top-1'': fraction of images retrieved correctly among the 1000 images of the other space; ``Same id.'': another image of the same identity (other space, or within one space); ``Diff.\ id.'': an image of a different identity in the other space.}
\label{lcmem:tab:cross_space}
\end{table}

\subsection{Extension to New Autoencoders}
\label{lcmem:sec:new_autoencoders}
To measure how fast a new autoencoder can be added, we hold out FLUX.2 (32 latent channels, no adapter in the model) and train only a new adapter ($0.02$M parameters) on FLUX.2 latents of the training images.
The target of each image is its atlas embedding (mean over the three training autoencoders), the loss is the cosine distance, and we use clean latents.
The backbone of the model is frozen. 
\Cref{lcmem:tab:alignment_curve} shows the validation \AUCROC{} over training.
The untrained adapter already passes some information: re-identification within FLUX.2 is above chance at step 0 ($0.696$).
After 1000 steps the gap to the trained autoencoders is closed ($0.964$ against $0.963$ to $0.965$).
The cosine to the atlas target is similar: a random other image has a cosine of $0.02$, a trained space reaches $0.82$, and the new adapter reaches $0.81$ after 1000 steps and $0.825$ at the end.
The final cosine is slightly above that of the trained spaces, because the adapter is trained on the mean target itself, whereas the student was trained on the target of each individual space.
With precomputed latents, alignment to new autoencoders takes less than one hour. 
All rows of \cref{lcmem:tab:alignment_curve} are evaluated on the same 5000 validation pairs. The row ``noisy latents'' trains the adapter with the latent noise of the student training (noise level per sample, clean with probability $0.2$); validation stays on clean latents.

\subsection{Recall at Higher Sampling Budgets}
\label{lcmem:sec:ova_budget}
\begin{figure*}[t]
    \centering
    \includegraphics[width=0.85\linewidth]{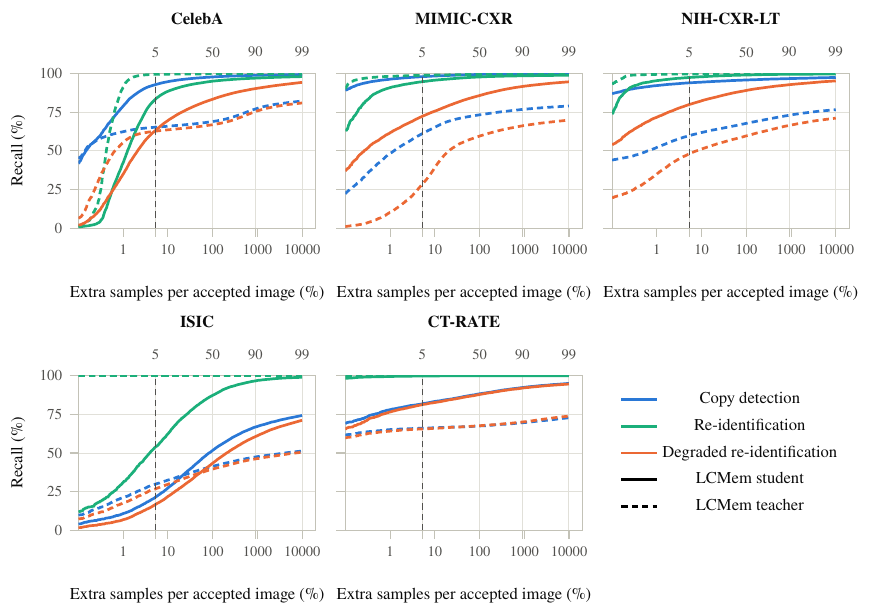}
    \caption{Recall against the sampling cost in the deployment case (training split, full atlas), one panel per dataset. Bottom axis: extra samples per accepted image, $\rho/(1-\rho)$ (log scale); top axis: rejection share $\rho$ of the novel queries (\%). Solid: LCMem student, dashed: LCMem teacher. Copy detection and degraded re-identification at corruption strength 4. Vertical dashed line: $\rho = 5\,\%$, the operating point of the tables.}
    \label{lcmem:fig:ova_budget}
\end{figure*}

\Cref{lcmem:fig:ova_budget} extends the one-vs-all deployment experiment (\cref{lcmem:sec:ova_deployment}) to larger rejection shares on the training split with the full atlas.
For every $\rho$, the threshold is the $(1-\rho)$-quantile of the one-vs-all scores of the novel queries of the same split, and the recall is the fraction of positive queries above it.
The bottom axis gives the price as extra samples per accepted image, $\rho/(1-\rho)$ (an extra cost of $100\,\%$ means twice the samples in total); the top axis gives the corresponding rejection share $\rho$.
\Cref{lcmem:tab:ova_budget} lists the recall at $\rho = 90\,\%$ (ten samples per accepted image).

The curves flatten at high budgets: on MIMIC, NIH-CXR-LT and CelebA, the recall of the student is already close to its final value at $100\,\%$ extra cost, and additional samples add only a few points.
The teacher re-identifies the training identities almost perfectly ($99.8\,\%$ macro) at this budget, but is clearly worse in copy detection and degraded re-identification ($68.7$ and $64.8\,\%$ macro).
On ISIC the teacher's re-identification is perfect on the training split because it has learned the identities of the training patients; on unseen test identities it is at chance (\cref{lcmem:sec:ova_deployment}), so this number measures recall of the training identities, not generalization.
No finite budget gives a recall of exactly $100\,\%$: a small fraction of positive queries scores below all novel queries (the lowest positive needs $\rho \approx 100\,\%$ in almost all settings).

\begin{table}[t]
\centering
\renewcommand{\arraystretch}{1.1}
\small
\setlength{\tabcolsep}{4pt}
\resizebox{\linewidth}{!}{%
\begin{tabular}{lccccccc}
\toprule
\rowcolor{fau}
\color{white} & \multicolumn{3}{c}{\color{white}\textbf{LCMem student}} & & \multicolumn{3}{c}{\color{white}\textbf{LCMem teacher}} \\
\rowcolor{fau}
\color{white}\textbf{Dataset} & \color{white}\textit{Copy} & \color{white}\textit{Re-ID} & \color{white}\textit{Deg.\ Re-ID} & & \color{white}\textit{Copy} & \color{white}\textit{Re-ID} & \color{white}\textit{Deg.\ Re-ID} \\
\midrule
CelebA & 98.8 & 96.9 & 90.2 & & 77.0 & 99.8 & 75.3 \\
MIMIC-CXR & 99.7 & 98.5 & 91.5 & & 76.7 & 99.6 & 66.1 \\
NIH-CXR-LT & 96.6 & 99.4 & 92.7 & & 72.9 & 99.9 & 66.3 \\
ISIC & 66.8 & 96.6 & 60.6 & & 47.4 & 100.0 & 46.2 \\
CT-RATE & 92.3 & 100.0 & 92.0 & & 69.6 & 100.0 & 70.1 \\
\midrule
Macro & 90.8 & 98.3 & 85.4 & & 68.7 & 99.8 & 64.8 \\
\bottomrule
\end{tabular}}
\caption{Recall (\%) at a rejection share $\rho = 90\,\%$ (ten samples per accepted image) in the one-vs-all deployment case (training split, full atlas; copy detection and degraded re-identification at corruption strength 4).}
\label{lcmem:tab:ova_budget}
\end{table}

\subsection{Implementation Details}
\label{lcmem:sec:impl_details}
This section gives the details of the network and of the two training stages that are omitted in \cref{lcmem:sec:network,lcmem:sec:teacher,lcmem:sec:student}.

\noindent\textbf{Architecture.}
The latent $z^{(j)} = \enc_j(x)$ of autoencoder $j$ is multiplied by a scalar $\kappa_j$, one constant per autoencoder measured on the training latents so that all spaces enter with the same scale, and passed through the adapter $A_j$.
The adapter has two parallel paths on the $64{\times}64$ grid that both read the scaled latent and are summed: a $1{\times}1$ convolution to $d_0 = 32$ channels, and a block of $3{\times}3$ convolution to $32$ channels, GroupNorm with 8 groups, GELU and a second $3{\times}3$ convolution ($0.036$M parameters for all three adapters).
The adapter output enters a custom ConvNeXt-Tiny~\cite{liu2022convnext} whose $4{\times}4$ stride-4 patchify stem is replaced by a $1{\times}1$ convolution with stride 1 from $32$ to $96$ channels, so that the first stage runs on the full $64{\times}64$ latent grid and the four stages see $64^2$, $32^2$, $16^2$ and $8^2$ positions.
The stride matters: a stem with stride 4 or stride 2 lowers the re-identification \AUCROC{} from $0.909$ to $0.858$ and $0.880$ (\cref{lcmem:tab:ablation}).
After global average pooling and the LayerNorm of the ConvNeXt classifier, the $768$-dimensional feature is mapped by a linear layer to $512$ dimensions and followed by a BatchNorm layer without further nonlinearity, which yields the embedding $u$.
Stochastic depth with rate $0.3$ is used in the ConvNeXt during training to avoid overfitting.
The custom ConvNeXt and the adapters are initialised by feature distillation from a ConvNeXt-Large on ImageNet. 

\noindent\textbf{Teacher: sampling.}
Each training sample is a pair $(x, x')$ drawn as follows: a dataset uniformly from $\mathcal{D}$ (so every dataset contributes one sixth of the samples regardless of its size), an identity of that dataset, and one sample $x$ of this dataset.
With probability $0.5$ the pair is positive and $x'$ is drawn from the same identity (with replacement, so $x' = x$ is possible and is the only positive for datasets with one image per identity). Otherwise it is negative and $x'$ is drawn from another identity of the same dataset, so negatives never cross datasets.
The autoencoder index $j$ is drawn independently and uniformly from $\{$SD2, SDXL, FLUX.1$\}$ for $x$ and for $x'$, so two thirds of the pairs mix latent spaces.
Each batch holds $B = 256$ pairs, i.e.\ $2B = 512$ embeddings $u_1, \dots, u_{2B}$, every one with its dataset, identity and image index.

\noindent\textbf{Teacher: contrastive term.}
$\mathcal{L}_{\mathrm{nce}}$ is the NT-Xent loss~\cite{chen2020simpleframework} over the positive pairs of the batch.
For the $B_+$ positive pairs this gives $2B_+$ embeddings; for each $u_i$, $k_+$ is the embedding of its pair partner and the remaining $2B_+ - 2$ embeddings of the other positive pairs are negatives,
\begin{equation}
\mathcal{L}_{\mathrm{nce}}
= - \frac{1}{2B_+} \sum_{i=1}^{2B_+} \log
\frac{\exp\left( \mathrm{sim}(u_i, k_{+}) / \tau \right)}
{\sum_{k \neq u_i} \exp\left( \mathrm{sim}(u_i, k) / \tau \right)} ,
\end{equation}
with temperature $\tau = 0.07$.
Because identities are drawn independently for each pair, a few of these negatives can share the anchor's identity by chance.%
The sampled negative pairs do not enter this term; they contribute their two images to the other two losses only.
This is the only term that compares two embeddings directly.

\noindent\textbf{Teacher: atlas term.}
After the loss, the atlas entry $\bar{u}_i$ is updated with momentum $\eta = 0.9$,
and on the first visit of an image the entry is set to $u_i / \lVert u_i \rVert$ without a loss.

\noindent\textbf{Teacher: identity term.}
For every dataset $d \in \mathcal{D}_{\mathrm{id}}$ with identity labels (all but ImageNet) we keep a separate ArcFace head~\cite{deng2019arcface}: a matrix $W_d \in \mathbb{R}^{K_d \times 512}$ with one learned prototype per training identity of $d$.
$\mathcal{L}_{\mathrm{arc}}$ is the standard ArcFace loss (scale $s = 64$, additive angular margin $m$) with the identity $\idl(x)$ as label, averaged over the batch embeddings of each dataset and summed over $d \in \mathcal{D}_{\mathrm{id}}$.
The softmax runs over the identities of one dataset only, so identities of different datasets never compete, and both images of every pair, positive or negative, enter with their identity.
The margin is ramped linearly from $0$ to $m = 0.5$ over the first $20{,}000$ steps.
The prototypes ($36.3$M parameters, dominated by MIMIC-CXR and NIH-CXR-LT) are used for training only and discarded afterwards.
Unlike the atlas entries, which are written from the embeddings, they change only through their gradient, and the softmax pushes $u$ away from all other prototypes of its dataset.

\noindent\textbf{Teacher: optimisation.}
AdamW with learning rate $10^{-4}$ and weight decay $0.05$ for the network, learning rate $10^{-3}$ and no weight decay for the ArcFace prototypes, bfloat16 autocast, at most $200{,}000$ steps.
Every $2{,}000$ one-vs-all validation scores are computed and training stops with patience of 10 after no improvements, which happened after roughly $7.7$\,hours on one H100. 

\noindent\textbf{Student: noise-level conditioning.}
At diffusion model training time, a clean latent $z$ is corrupted with Gaussian noise of standard deviation $\sigma$, $z + \sigma\epsilon$ with $\epsilon \sim \mathcal{N}(0, \mathbf{I})$ drawn independently of $z$, and every state of the sampler is such a mixture of signal and noise, fully described by its signal-to-noise ratio $\mathrm{SNR} = 1/\sigma^2$ (a state $\alpha_t z + s_t \epsilon$ of any schedule is the same up to the scalar $\alpha_t$, with $\sigma_t = s_t / \alpha_t$).
$\log\sigma / 4$ is mapped to $256$ Fourier features (random frequencies and phases, fixed), passed through two magnitude-preserving linear layers with SiLU as in EDM2~\cite{karras_analyzing_2024}, and projected by one zero-initialised linear layer per stage to a per-channel scale $\gamma$ and shift $\beta$, applied after each of the four ConvNeXt stages as $h \leftarrow h (1 + \gamma) + \beta$.

\noindent\textbf{Student: training input.}
Each training image is augmented in pixel space with the strong set used for the baselines: random affine rotation of $\pm 20^\circ$ with probability $0.5$, horizontal flip with probability $0.5$, Gaussian blur with standard deviation up to $0.2$, Gaussian noise with standard deviation up to $0.05$, clamping to $[0,1]$, and JPEG compression with quality uniform in $[10, 95]$ with probability $0.5$.
An autoencoder $j$ is drawn uniformly per image and the augmented image is encoded online by the frozen $\enc_j$ (posterior mean, full precision).
The noise is added before the adapter, and $\sigma$ enters the FiLM layers as conditioning.

\noindent\textbf{Student: target and optimisation.}
Before training the student, the teacher embeds every training image once, clean and un-augmented, in each of the three latent spaces, $u^{(j)}(x) = S_\phi(A_j(\kappa_j \enc_j(x)))$; these embeddings are stored (the deployment atlas of the teacher, half precision).
Adam with learning rate $10^{-4}$, no weight decay, batches of $32$ images (the online encoding by the autoencoder dominates the cost), bfloat16 autocast for the network.
We use the same validation procedure as for the teacher.

\subsection{Baseline Details}
\label{lcmem:sec:baseline_details}
\emph{Pixel-wise metrics.} The mean squared error (MSE), the mean absolute error (L1), the normalized cross-correlation (NCC) and \ac{SSIM} \cite{wang_image_2004} compare the two images directly at $512 \times 512$.

\emph{Pretrained encoders.} Following the selection of \cite{stein2024exposing, dombrowski_cvpr}, we use the features of BYOL \cite{grill_bootstrap_2020}, CLIP \cite{radford2021learning}, ConvNeXt \cite{liu2022convnext}, data2vec \cite{baevski_data2vec_2022}, DINOv2 \cite{oquab_dinov2_2024}, Inception \cite{szegedy_rethinking_2016}, MAE \cite{he_masked_2022} and SwAV \cite{caron_unsupervised_2020}, extended with DINOv3 \cite{siméoni2025dinov3} and the VGG19 \cite{simonyan2014very} features of \cite{shi2026privacypreservinglatent}, and score a pair by the cosine similarity of its features.
For DINOv3 we only consider the output of the pooling layer of the smallest model, because the computational demands of the larger models make their use infeasible for many low- to mid-resource tasks such as online evaluation of memorization.
Preliminary results on smaller subsets showed no difference between the larger and the smaller models.
The ``Random'' baseline is an untrained Inception-v3 \cite{szegedy_rethinking_2016}, which has nevertheless been shown to provide useful feature representations \cite{oareilly_pre-trained_2021}.

\emph{Copy detection.} SSCD~\cite{pizzi2022selfsuperviseddescriptor} is a self-supervised descriptor trained for image copy detection and a common memorization detector for diffusion models~\cite{somepalli2023understandingmitigating}.
pHash~\cite{zaunerimplementationbenchmarking} is a $64$-bit perceptual hash that is compared by its Hamming distance.

\emph{Trained on our data.} These baselines are trained on the same training data of LCMem, on $512 \times 512$ images, with early stopping on the validation \AUCROC{}.
Unless stated otherwise the backbone is an ImageNet-pretrained ResNet-101 with $1024$ output features, trained with batch size $64$ and learning rate $10^{-4}$.
\emph{Siamese} is the re-identification model of \cite{packhäuser2022deeplearning}, a binary cross-entropy on a linear head over the absolute feature difference of a pair.
\emph{Siamese + NT-Xent} is the contrastive model of \cite{dar2024unconditional} with the identity labels as positives; \emph{Siamese + NT-Xent (unsup.)} is their original setting without identity labels, where the positive of an image is its own augmented copy.
\emph{ArcFace}~\cite{deng2019arcface} classifies the training identities with an additive angular margin ($s = 64$, $m = 0.5$, batch size $128$) and scores a pair by the cosine of its embeddings; \emph{ArcFace (ConvNeXt-T)} replaces the ResNet-101 by a ConvNeXt-Tiny and is the image-space counterpart of the LCMem backbone.
ArcFace needs identity labels, so the ArcFace models are trained on the five datasets of \cref{lcmem:tab:dataset_overview} without ImageNet; all other trained baselines and LCMem also use the ImageNet subset.
The original methods use weak augmentation (Gaussian noise for Siamese; rotation by up to $5^\circ$ and horizontal flip for the others).
The rows marked \emph{strong aug.} are trained with the pixel augmentation of the LCMem student instead (\cref{lcmem:sec:student}) and are the augmentation-matched control.
The threshold of every baseline is fitted with the same protocol as for LCMem.

\subsection{Generator of the Applied Experiment}
\label{lcmem:sec:generator}
We use EDM2~\cite{karras_analyzing_2024} in the FLUX.2 latent space, $z = \enc(x) \in \mathbb R^{64\times64\times32}$ for a $512\times512$ image, with every latent channel normalized to zero mean and a standard deviation of 0.5 over the training set. 
The denoiser $D_\theta(z_t;\sigma_t,c)$ is an EDM2-M network that was pretrained on ImageNet in the same latent space and fine-tuned on the training images of CelebA (batch size 1024, reference learning rate 0.01; we use the exponential moving average of the weights after 50M training images). 
It is conditioned on the 40 binary attributes $c\in\{0,1\}^{40}$ of CelebA, which are normalized to unit length and embedded by a two-layer MLP. 
Noisy latents are $z_t = z + \sigma_t\epsilon$ with $\epsilon\sim\mathcal N(0,\mathbf I)$, i.e. $\alpha_t = 1$ and the noise level is $\sigma=\sigma_t$. 
We sample with the deterministic second-order Heun sampler of EDM2 with $T=32$ steps on its noise schedule, from $z_T=\sigma_T\epsilon$ with $\sigma_T=80$ down to $\sigma_t=0.002$. 
The conditional denoiser is guided by an unconditional denoiser $D_{\theta'}$ with guidance scale $w=1.1$.

\end{document}